\documentclass[11pt]{article}

\usepackage[a4paper,margin=1in]{geometry}

\usepackage[utf8]{inputenc}
\usepackage[T1]{fontenc}

\usepackage{fvextra}      
\usepackage{longtable}
\usepackage{pdflscape}

\usepackage{tabularx}
\usepackage{array}

\usepackage{microtype}
\usepackage{setspace}

\usepackage{amsmath}
\usepackage{mathtools}
\usepackage{bm}

\usepackage{newpxtext}
\usepackage{newpxmath}

\newcommand{\AynMark}{%
  \raisebox{0.15ex}{\rotatebox[origin=c]{180}{\textquotesingle}}%
}
\newcommand{\HamzaMark}{%
  \raisebox{0.15ex}{\textquotesingle}%
}

\DeclareUnicodeCharacter{02BF}{\AynMark}   
\DeclareUnicodeCharacter{02BE}{\HamzaMark} 

\usepackage{graphicx}
\usepackage{booktabs}
\usepackage{multirow}
\usepackage{array}
\usepackage{tabularx}
\usepackage{threeparttable}
\usepackage{caption}
\usepackage{subcaption}

\usepackage{longtable}
\usepackage{pdflscape}
\usepackage{fvextra}
\usepackage{placeins}

\usepackage{enumitem}
\setlist[itemize]{leftmargin=1.4em,itemsep=0.2em,topsep=0.3em}
\setlist[enumerate]{leftmargin=1.6em,itemsep=0.2em,topsep=0.3em}

\usepackage{xcolor}
\definecolor{mherblue}{HTML}{2F4B67}
\definecolor{mhergray}{HTML}{5E6470}
\definecolor{linkblue}{HTML}{315F85}

\usepackage{titlesec}

\titleformat{\section}
  {\Large\bfseries\color{mherblue}}
  {\thesection}
  {0.7em}
  {}

\titleformat{\subsection}
  {\large\bfseries\color{mherblue}}
  {\thesubsection}
  {0.7em}
  {}

\titleformat{\subsubsection}
  {\normalsize\bfseries\color{mhergray}}
  {\thesubsubsection}
  {0.7em}
  {}

\titlespacing*{\section}{0pt}{2.0em}{0.7em}
\titlespacing*{\subsection}{0pt}{1.4em}{0.5em}

\usepackage[round,authoryear]{natbib}

\usepackage[
    colorlinks=true,
    linkcolor=linkblue,
    citecolor=linkblue,
    urlcolor=linkblue
]{hyperref}

\usepackage[nameinlink,capitalise,noabbrev]{cleveref}

\usepackage{siunitx}
\usepackage{url}
\usepackage{xspace}

\title{
    \vspace{-1.2cm}
    \textbf{
    When Names Cross Scripts:
    A Source-Grounded Benchmark for Historical Entity Reconciliation
    in the Mongol World
    }
}

\author{%
\begin{tabular}{c@{\hspace{4em}}c}
Xiang Chen$^{1,*}$ &
Zeyu Zhang$^{2}$ \\[0.45em]
{\small $^{1}$Independent Researcher} &
{\small $^{2}$University of Amsterdam \& Amsterdam UMC} \\[0.2em]
{\small \texttt{chenxiang001a@gmail.com}} &
{\small \texttt{z.zhang2@uva.nl}} \\[0.35em]
\multicolumn{2}{c}{\small $^{*}$Corresponding author}
\end{tabular}%
}

\date{}

\begin{document}

\maketitle

\vspace{-0.5em}

\begin{abstract}

Historical people may appear under different languages, scripts, and
transcription traditions, while distinct individuals may share highly similar
or even identical names. This makes historical identity reconciliation more
than a problem of string matching or transliteration. We introduce
\textbf{MHER}, a provenance-controlled benchmark for pairwise reconciliation
of person-name attestations from the Mongol world. MHER contains a balanced
396-pair Name-only core over 84 primary historical persons and a stricter
160-pair Source-grounded subset constructed from mention$\times$source
evidence, with entity-disjoint development and test splits.

Across five generative systems, correctly Source-grounded evidence improves
paired TEST accuracy by 12.96--94.44 percentage points relative to Name-only
input. On five identical-surface \textsc{different}-person cases, all models
fail under names alone (0/25 model--item decisions), whereas Source-grounded
evidence yields 24/25 correct resolutions, with the remaining output an
abstention. Context-only ablations show that historical descriptions often
carry substantial identity information, while explicitly signaled misgrounding controls produce substantially lower performance. We also find that names are
not uniformly beneficial: for Qwen3-8B, restoring surface forms converts ten
otherwise correct Context-only distinctions into false identity merges.

These results show that historical entity reconciliation depends not only on
surface correspondence, but on whether identity judgments respond
appropriately to provenance-controlled historical evidence. MHER therefore
provides a controlled framework for studying evidence use, abstention, and
failure modes in historical NLP.

\end{abstract}

\vspace{0.5em}


\section{Introduction}
\label{sec:introduction}

Historical people rarely possess a single stable name across the records that
preserve them. The same individual may appear under forms shaped by different
languages, scripts, transcription systems, scribal conventions, titles, and
later scholarly romanizations. Conversely, distinct individuals may share
near-identical---or even exactly identical---name forms. Before records from
different sources can be aggregated into a common historical account, one
must therefore answer a deceptively simple question:

\begin{quote}
\emph{Do these two attestations refer to the same historical person?}
\end{quote}

This question cannot in general be reduced to name matching. Cross-script
normalization and transliteration can make orthographically distant forms more
comparable, but surface correspondence is only one source of identity
evidence. Two distant forms may denote the same person, while two identical
forms may denote different people. In MHER, for example, cross-tradition
attestations of the same ruler can have little direct surface overlap, whereas
two attestations written identically as \emph{Oghul Qaimish} correspond to
different historical individuals whose identities are distinguished only by
their chronology, kinship, and political setting. Such cases make the
difference between matching a name and reconciling a person explicit.

Modern entity linking typically addresses a neighboring but structurally
different problem: mapping a textual mention to a canonical entity in a
predefined knowledge base. Multilingual systems have extended this paradigm
across large numbers of languages and scripts, while cross-language entity
clustering and record linkage have long shown that identity relations can also
be inferred directly between observations without assuming that a complete
entity inventory is known \citep{green2012entity,fellegi1969theory}.
Historical settings make this distinction particularly consequential.
Knowledge bases may omit long-tail individuals, documentary evidence may be
fragmentary or inconsistent, and the reconciliation represented in a modern
reference resource may itself be the outcome of historical scholarship rather
than an uncontested starting point.

Recent historical NLP resources have substantially broadened entity
processing beyond contemporary, high-resource settings. Historical Chinese
newspapers, long-tail historical knowledge extraction, Sanskrit literary
entities, and multilingual historical entity linking all expose difficulties
arising from spelling variation, diachrony, sparse knowledge-base coverage,
and ambiguous person names
\citep{blouin2024dataset,graciotti2025kemhisto,
sarkar2025mahanama,santini2026confidence}.
Historical record linkage reaches a complementary conclusion: names are
non-unique and error-prone, and identity decisions often become more reliable
when biographical attributes such as family relationships, geography, age, or
other contextual information are considered together
\citep{abramitzky2020automated}. These traditions suggest that historical
identity should be treated as an evidence-combination problem rather than as
surface-form similarity alone.

Yet providing ``context'' creates its own evaluation problem. A historical
description is useful evidence only if it is actually associated with the
attestation whose identity is being judged. Context copied from a canonical
entity description may accidentally reveal the answer; two \textsc{same}
mentions may share duplicated material; or generic topical similarity may
become an unintended label cue. A contextual benchmark can therefore appear
to measure historical reasoning while in fact rewarding artifacts of how the
context was constructed. For historical entity reconciliation, provenance is
consequently not only documentation about an example: it is part of what
makes the example semantically valid.

We introduce \textbf{MHER} (\emph{Mongol-world Historical Entity
Reconciliation}), a provenance-controlled benchmark for studying this
problem across languages, scripts, and transcription traditions. Rather than
requiring each mention to resolve first to a modern knowledge-base identifier,
MHER asks directly whether two source-attested person-name mentions denote the
same historical individual. The primary gold relation is binary
(\textsc{same}/\textsc{different}), while evaluated systems may explicitly
abstain when the evidence presented to them is insufficient.

MHER separates two questions that are often conflated. The first is what can
be inferred from the name forms themselves. The second is what becomes
recoverable when independently source-grounded historical evidence is made
available. To make this distinction measurable, the benchmark pairs
Name-only evaluation with a stricter Source-grounded subset in which evidence
is associated at the mention$\times$source level. Context-only ablation and
deterministically mismatched context controls further distinguish the
information contained in historical descriptions from the correctness of
their grounding. The resulting framework is intended not as another
large-scale name resource, but as a controlled setting in which surface
evidence, historical evidence, and evidence provenance can be varied
separately.

The Mongol world provides a particularly suitable testbed for this problem.
Historical persons are preserved across heterogeneous linguistic and
documentary traditions, with names represented through different scripts,
transcription conventions, titles, and later scholarly practices. At the same
time, the benchmark is deliberately narrow in its historical claims: MHER is
not intended to establish a universal solution to historical entity
resolution. Its purpose is to make one recurring historical NLP problem
experimentally tractable while retaining the source relationships on which
the identity judgments depend.

Our contributions are fourfold:

\begin{itemize}

    \item \textbf{Historical entity reconciliation as an evidence-conditioned
    task.}
    We operationalize pairwise identity reconciliation between historical
    person-name attestations, distinguishing the task from conventional
    mention-to-KB linking and from surface-form correspondence, while allowing
    explicit abstention when a binary judgment is not warranted.

    \item \textbf{A provenance-controlled benchmark.}
    We introduce MHER, comprising a balanced 396-pair Name-only core over
    84 primary historical persons with entity-disjoint DEV/TEST splits, and a
    stricter 160-pair Source-grounded subset constructed from
    mention$\times$source-specific evidence.

    \item \textbf{Leakage-aware evidence intervention and falsification.}
    We treat source provenance as part of benchmark construction, require
    independently grounded evidence for contextual \textsc{same} pairs, and
    accompany the paired Name-only/Source-grounded intervention with
    Context-only and three deterministic shuffled-context controls designed to
    separate correct grounding from the generic presence of additional text.

    \item \textbf{Evaluation beyond forced binary accuracy.}
    We combine lexical, transliteration, multilingual embedding, proprietary,
    and open-weight systems with explicit abstention, probabilistic scoring,
    surface-confusable challenge cases, repeated-entity sensitivity, and a
    separate expert-unresolved challenge. This evaluation framework is
    designed to distinguish incorrect identity resolution from uncertainty,
    surface-form reliance, and failure to use the supplied historical evidence.

\end{itemize}

The broader premise of this work is simple: \emph{names provide evidence about
identity, but they are not themselves the identity relation}. Historical NLP
systems should therefore be evaluated not only on whether they can recognize
a familiar name, but on whether their identity judgments respond
appropriately to the source-grounded evidence available for that person.

\section{Related Work}
\label{sec:related-work}

MHER lies at the intersection of multilingual entity linking, cross-script
name processing, historical NLP, record linkage, and evidence-grounded model
evaluation. None of these components is individually new. The distinction of
the present setting is their combination: direct pairwise reconciliation of
historical attestations, mention-level source provenance, and a controlled
intervention that separates name-form evidence from correctly and incorrectly
grounded historical context.

\subsection{Multilingual Entity Linking and Pairwise Identity Resolution}
\label{sec:rw-entity-linking}

Entity linking conventionally maps a textual mention to a canonical entity in
a predefined knowledge base. Cross-lingual and multilingual work has extended
this paradigm to increasingly broad language coverage. \citet{pan2017crosslingual}
develop cross-lingual name tagging and linking for 282 languages, while
\citet{botha2020entity} formulate multilingual entity linking against a
language-agnostic KB covering more than 100 languages and millions of
entities. Rare entities and low-resource languages remain particularly
difficult in this setting. mGENRE instead formulates multilingual entity
linking autoregressively, generating multilingual entity names while still
resolving mentions to a KB inventory \citep{decao2022multilingual}.

Mention-centric formulations bring this literature closer to MHER.
MOLEMAN learns contextual representations of multilingual entity mentions
and retrieves labeled mentions of the same entity rather than relying on a
single entity vector \citep{fitzgerald2021moleman}. Nevertheless, its mention
pairs ultimately inherit entity identities from a KB-linked training corpus.

Pairwise identity inference without assuming that the complete entity
inventory is observed also predates MHER. Most directly,
\citet{green2012entity} study cross-language entity clustering across
documents, inferring which mentions refer to the same entity without assuming
that the true entity inventory is known in advance. This is an important
precedent for the structural distinction we make between

\begin{equation}
    m \rightarrow e_{\mathrm{KB}}
\end{equation}

and

\begin{equation}
    m_a \stackrel{?}{\equiv} m_b .
\end{equation}

MHER therefore does not claim pairwise or KB-independent identity inference as
a new task family. Its focus is narrower: historically attested person
mentions for which identity evidence itself may have to be reconstructed
across linguistic, script, and documentary traditions.

\subsection{Cross-Script Names and Transliteration}
\label{sec:rw-transliteration}

Name variation across writing systems has a long history in multilingual NLP.
Transliteration is frequently used to increase correspondence between names
written in different scripts and can improve candidate generation for
cross-lingual entity linking. \citet{upadhyay2018bootstrapping}, for example,
develop low-resource transliteration methods and evaluate their contribution
to cross-lingual EL candidate generation. \citet{khakhmovich2020crosslingual}
study multilingual personal-name transliteration and cross-lingual named
entity list search across a large number of languages and scripts.

Recent resources have substantially expanded the scale of multilingual name
data. ParaNames contains approximately 140 million names for 16.8 million
entities across more than 400 languages and is explicitly intended to support
tasks including name translation, transliteration, NER, and entity linking
\citep{saleva2024paranames}. A recent survey characterizes differences in
writing systems as a persistent ``script barrier'' in cross-lingual NLP and
reviews the benefits and information trade-offs associated with
transliteration \citep{jayakumar2026scripts}.

These works motivate the transliteration-aware baselines in MHER, but name
correspondence and historical identity remain different relations.
Transliteration can help determine whether two strings are plausible
cross-script renderings, but it cannot determine whether two people who share
an identical or highly similar name are historically the same individual.
Conversely, historically equivalent attestations may remain orthographically
distant after simple transliteration. MHER therefore treats surface
correspondence as one source of evidence rather than as the target relation
itself.

\subsection{Historical NLP, Entity Linking, and Record Linkage}
\label{sec:rw-historical}

Historical text processing introduces a combination of difficulties that are
less severe in many contemporary NLP settings, including OCR or transcription
noise, diachronic language variation, heterogeneous documentary conventions,
and sparse resources \citep{ehrmann2023historicalner}. Recent work has begun
to move historical entity processing beyond NER toward linking, coreference,
and structured historical knowledge.

\citet{blouin2024dataset} introduce a dataset based on Chinese historical
newspapers from 1872--1949 covering NER, entity linking, coreference, and
entity relations, demonstrating that historical entity processing extends
well beyond modern European and Latin-script corpora. KE-MHISTO focuses on
multilingual historical knowledge extraction and explicitly targets the
long-tail problem, where contemporary language models and knowledge resources
provide uneven coverage of historically obscure entities
\citep{graciotti2025kemhisto}.

Mah\={a}n\={a}ma provides an especially close neighboring setting
\citep{sarkar2025mahanama}. Built from the Sanskrit
\emph{Mah\={a}bh\={a}rata}, it contains more than 109,000 mentions linked to
approximately 5,500 entities and exhibits extensive name variation,
ambiguity, and long-range contextual dependencies. It demonstrates that
historical and literary entity resolution can require considerably more than
robust surface matching.

The closest recent system-level work is MHEL-LLaMo
\citep{santini2026confidence}, which applies multilingual bi-encoder retrieval
and LLM-based confidence-aware candidate selection to historical entity
linking in six European languages. This work is particularly important for
positioning MHER: multilingual historical EL with LLMs already exists, and
MHER is not the first benchmark or system in that category. The task boundary
is instead structural. MHEL-LLaMo remains a mention-to-candidate/KB linking
problem, including NIL prediction, whereas MHER asks directly whether two
source attestations denote the same historical person and manipulates the
evidence attached independently to the two mentions.

A complementary tradition comes from statistical record linkage.
The classical Fellegi--Sunter framework treats identity matching as inference
from multiple partially informative record fields
\citep{fellegi1969theory}. Historical record linkage has subsequently shown
that person matching is complicated by non-unique names, spelling variation,
measurement error, and sparse records, and that supplementary information such
as geography and family relationships can materially improve linkage
\citep{abramitzky2020automated}. MHER shares this evidence-combination view,
but studies source-attested multilingual historical names and model-visible
narrative evidence rather than tabular census-style records.

\subsection{Source Grounding, Falsification, and Abstention}
\label{sec:rw-grounding}

The distinction between having additional text and being correctly grounded in
evidence is also central to contemporary language-model research.
Retrieval-augmented generation conditions language models on retrieved
documents in order to improve knowledge-intensive prediction
\citep{lewis2020retrieval}. Subsequent work on attributed generation shifts
attention from the mere presence of retrieved material toward whether model
claims are supported by verifiable sources. ALCE evaluates language-model
answers together with their citations \citep{gao2023enabling}, while
ExpertQA studies factuality and attribution using expert-curated questions and
expert evaluation across specialized domains \citep{malaviya2024expertqa}.

MHER addresses a related issue at the \emph{input} rather than output level.
A context is valid historical evidence only when it is correctly associated
with the attestation under consideration. This motivates our
mention$\times$source representation and the shuffled-context controls, which
preserve the historical context pool while deliberately breaking its
alignment with benchmark mentions.

More generally, NLP benchmarks are known to support unintended shortcuts.
Annotation artifacts can make labels predictable from superficial cues
\citep{gururangan2018annotation}, and models may achieve high benchmark scores
by relying on heuristics that fail under controlled counterexamples
\citep{mccoy2019right}. Contrast-set evaluation consequently advocates
meaningful perturbations of existing instances to test whether model decision
boundaries reflect the intended capability \citep{gardner2020evaluating}.
MHER adopts the same challenge-set logic but uses explicitly signaled
context derangements as an informed stress test: the historical context
pool is preserved while pair-specific grounding is deliberately invalidated
and disclosed.

Finally, MHER permits systems to abstain rather than forcing every historical
identity question into a binary decision. Selective prediction formalizes the
choice between answering and abstaining in NLP \citep{xin2021art}, while
recent LLM work emphasizes abstention as a distinct capability requiring its
own evaluation \citep{wen2025know}. Entity linking has a related notion of
NIL prediction, where a mention has no appropriate target in the available KB
\citep{zhu2023nil}. MHER's \textsc{ambiguous} action is different from NIL:
both historical persons may exist and be well attested, while the evidence
visible for a particular reconciliation question may still be insufficient
to justify either \textsc{same} or \textsc{different}.

Taken together, prior work establishes each of the major ingredients behind
MHER: multilingual and cross-script entity processing, pairwise identity
inference, historical EL, contextual record linkage, evidence grounding,
benchmark falsification, and selective prediction. Our contribution is not to
claim priority over these individual traditions, but to combine them in a
provenance-controlled historical benchmark where the same identity question
can be evaluated under name-only, correctly grounded, context-only, and
deliberately misgrounded evidence.

\section{Historical Entity Reconciliation}
\label{sec:task}

Historical people rarely possess a single stable name across the records in which they appear. 
The same individual may be represented through different languages, scripts, transcription 
systems, scribal conventions, and later scholarly romanizations. Conversely, distinct individuals 
may share identical or near-identical name forms. The resulting computational problem is therefore 
not merely to determine whether two strings resemble one another, but whether two historically 
attested mentions denote the same person.

We call this task \emph{historical entity reconciliation}. Given two person-name mentions drawn 
from historical or scholarly sources, the system must determine whether they refer to the same 
underlying historical individual. The formulation is deliberately pairwise: unlike conventional 
entity linking, reconciliation does not require either mention to be mapped first to a complete, 
stable, and uncontested modern knowledge-base inventory.

\subsection{Task Formulation}
\label{sec:task-definition}

Let a historical name mention be represented as
\begin{equation}
    m_i = (s_i, \ell_i, \sigma_i),
\end{equation}
where $s_i$ is the attested or scholarly represented surface form, $\ell_i$ denotes its language, 
and $\sigma_i$ denotes its script, transcription system, or representation tradition. Given two 
mentions $(m_a,m_b)$, the underlying identity relation is

\begin{equation}
    y(m_a,m_b) \in 
    \{\textsc{same},\textsc{different}\}.
\end{equation}

A \textsc{same} relation states that the two mentions refer to the same historical person; 
\textsc{different} states that they refer to distinct individuals.

This relation differs structurally from conventional entity linking. In standard knowledge-base 
linking, a mention is resolved to an entity in a predefined inventory,

\begin{equation}
    m \rightarrow e_{\mathrm{KB}},
\end{equation}

whereas historical reconciliation directly evaluates

\begin{equation}
    m_a \stackrel{?}{\equiv} m_b.
\end{equation}

The distinction matters in historical settings because the target entity inventory may itself be 
incomplete. Long-tail individuals may have no modern knowledge-base entry, historical sources may 
preserve only partial information, and modern reference resources may encode one scholarly 
reconciliation among several possibilities. Pairwise reconciliation therefore asks a narrower 
question than full entity linking: whether the available evidence warrants treating two 
attestations as references to the same person.

\subsection{Name Correspondence Is Not Identity}
\label{sec:name-vs-identity}

A central premise of the task is that \emph{name correspondence and historical identity are not 
equivalent relations}. A name-matching system primarily estimates some form of correspondence 
between surface representations,

\begin{equation}
    \operatorname{sim}(s_a,s_b),
\end{equation}

possibly after normalization, transliteration, phonological approximation, or learned multilingual 
representation. Such correspondence is useful evidence for reconciliation, but it is neither 
necessary nor sufficient for identity.

It is not necessary because the same historical person may acquire substantially different surface 
forms across languages and documentary traditions. Script conversion is often non-bijective; 
transcriptions may reflect different phonological analyses; Chinese-character renderings, 
alphabetic transcriptions, and later romanizations need not preserve transparent character-level 
similarity; and titles, patronymics, lineage markers, or abbreviated forms may be inconsistently 
included.

Surface correspondence is also not sufficient. Historical naming systems frequently reuse personal 
names, titles, and lineage-associated forms. Two distinct individuals may therefore have highly 
similar names, or even exactly the same normalized surface representation. In such cases, increasing 
the sophistication of string normalization cannot by itself recover the identity relation: the 
relevant distinction lies outside the name string.

Formally, high surface similarity does not imply identity,

\begin{equation}
    \operatorname{sim}(s_a,s_b) \approx 1
    \;\not\Rightarrow\;
    y(m_a,m_b)=\textsc{same},
\end{equation}

and low similarity does not imply non-identity,

\begin{equation}
    \operatorname{sim}(s_a,s_b) \approx 0
    \;\not\Rightarrow\;
    y(m_a,m_b)=\textsc{different}.
\end{equation}

Historical entity reconciliation must therefore admit evidence beyond form correspondence.

\subsection{Evidence-Conditioned Reconciliation}
\label{sec:evidence-conditioned}

We distinguish between evidence intrinsic to the name representation and evidence supplied by the 
historical record. Let $e_i$ denote source-grounded evidence associated with mention $m_i$. Such 
evidence may describe chronology, kinship, lineage or group affiliation, political or military 
office, geographical association, or participation in historically documented events.

The reconciliation problem can then be written more generally as

\begin{equation}
    p\!\left(
        y \mid m_a,m_b,e_a,e_b
    \right).
\end{equation}

This formulation motivates two conceptually distinct evidence regimes.

In the \textbf{Name-only} regime, the system observes the two mention representations:

\begin{equation}
    x_i^{\mathrm{name}}=(m_a,m_b).
\end{equation}

Performance under this condition may reflect several signals simultaneously: orthographic or 
phonological correspondence, regularities learned across scripts and languages, prior familiarity 
with prominent historical figures, and the model's willingness to resolve or abstain when names 
alone are uncertain.

In the \textbf{Source-grounded} regime, each mention is additionally accompanied by evidence tied 
to the historical source associated with that mention:

\begin{equation}
    x_i^{\mathrm{source}}
    =
    \big((m_a,e_a),(m_b,e_b)\big).
\end{equation}

The underlying identity question is unchanged. What changes is the evidence made available for 
answering it. For paired items, the intervention

\begin{equation}
    x_i^{\mathrm{name}}
    \longrightarrow
    x_i^{\mathrm{source}}
\end{equation}

therefore asks whether historically grounded information changes a reconciliation decision that 
would otherwise have to be made from the names themselves.

Crucially, ``more context'' and ``historical evidence'' are not synonymous. Evidence is informative 
for reconciliation only insofar as it is correctly associated with the source attestations being 
compared. Establishing this association, and preventing contextual material from directly encoding 
the gold identity relation, is therefore part of benchmark construction rather than a purely 
presentational choice. We return to provenance and leakage control in Section~\ref{sec:benchmark}, 
and to falsification through alternative context conditions in Section~\ref{sec:falsification}.

\subsection{Resolution and Abstention}
\label{sec:abstention}

Historical evidence does not always justify a categorical identity decision. We therefore separate 
the benchmark's underlying identity relation from the actions available to an evaluated system.

For the primary benchmark, gold identity remains binary:

\begin{equation}
    \mathcal{Y}
    =
    \{\textsc{same},\textsc{different}\}.
\end{equation}

Models, however, may produce

\begin{equation}
    \mathcal{A}
    =
    \{\textsc{same},\textsc{different},\textsc{ambiguous}\},
\end{equation}

where \textsc{ambiguous} represents abstention rather than a third identity class. An abstaining 
system declines to assert either relation under the evidence presented to it.

This distinction is important because forced-choice evaluation conflates two qualitatively different 
failure modes: resolving a pair incorrectly and recognizing that the available evidence is 
insufficient for resolution. We therefore evaluate both ordinary reconciliation performance and 
selective behavior conditional on whether a model chooses to make a substantive 
\textsc{same}/\textsc{different} decision. Cases for which the underlying historical identification 
is itself treated as unresolved are evaluated separately rather than being folded into the binary 
benchmark core.

\subsection{Scope}
\label{sec:task-scope}

MHER instantiates historical entity reconciliation for personal names associated with the Mongol 
world and represented across heterogeneous linguistic, script, transcription, and scholarly 
traditions. The task begins from already identified person-name mentions and therefore does not 
evaluate named-entity recognition. Nor does it attempt to reconstruct a complete historical 
knowledge base.

Its central question is more specific: given two historical attestations, what can a computational 
system infer about their identity from the name forms alone, and what changes when independently 
source-grounded historical evidence becomes available?

This formulation makes name variation an important part of the problem without reducing historical 
identity to name matching. Names provide evidence about identity; they are not themselves the 
identity relation.

\section{The MHER Benchmark}
\label{sec:benchmark}

MHER is constructed from a source-linked registry of historical persons and
their attested or scholarly represented name forms. The benchmark is designed
so that historical identity is established before evaluation pairs are
created: pair membership does not itself serve as evidence for the underlying
identity relation, and model-visible inputs are separated from the scholarly
information used to construct the gold standard.

\begin{figure}[ht]
    \centering
    \includegraphics[width=\linewidth]
        {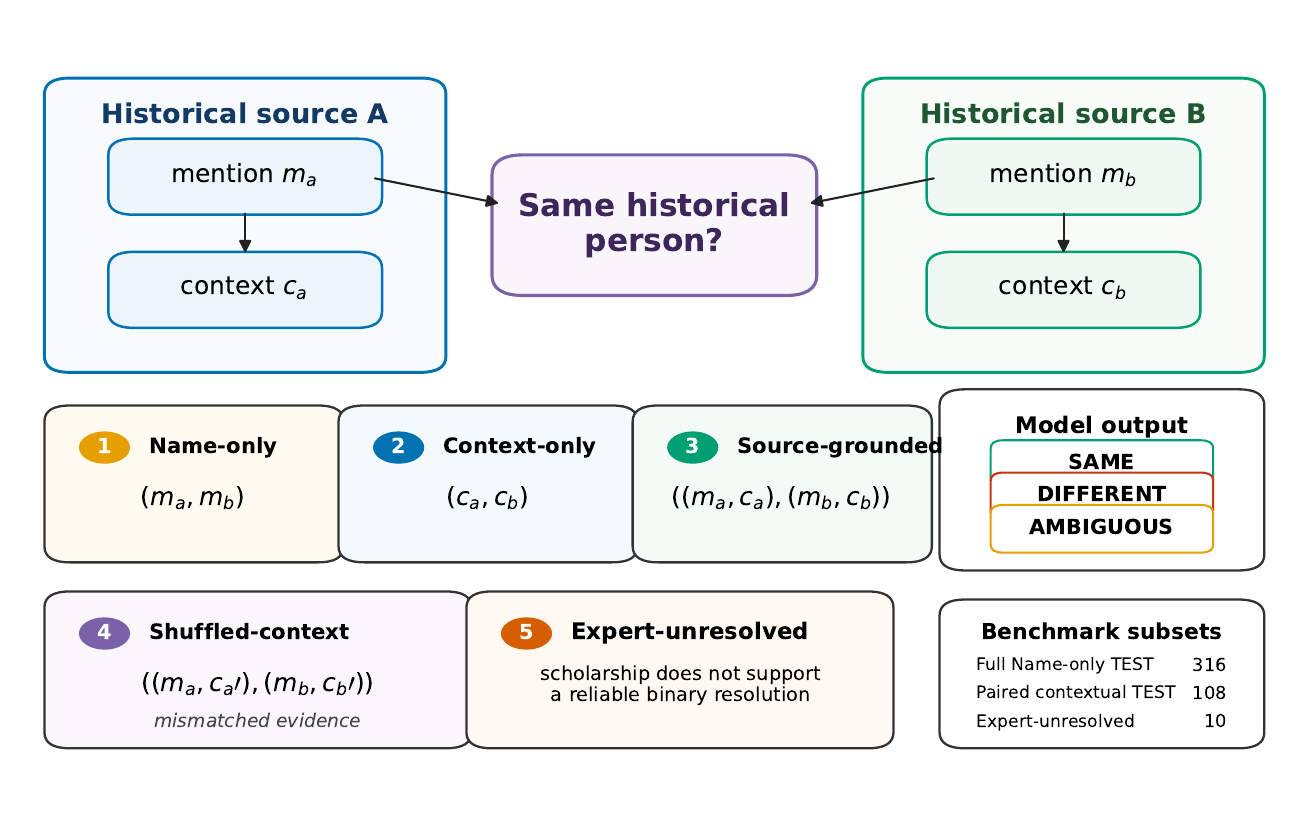}
    \caption{
    Overview of MHER benchmark construction and evaluation conditions.
    A source-linked scholarly registry defines canonical historical entities
    and their multilingual or multiscript mentions before evaluation pairs are
    generated. The balanced Name-only core is split by canonical entity.
    A stricter provenance-eligible subset is then constructed from
    mention$\times$source-specific evidence, yielding paired
    Source-grounded evaluation. Context-only and deterministic shuffled-context
    controls manipulate the same paired items without altering the underlying
    gold identity relation. Expert-unresolved cases are retained separately
    from the binary benchmark core.
    }
    \label{fig:benchmark-overview}
\end{figure}

Figure~\ref{fig:benchmark-overview} summarizes the benchmark pipeline, and
Table~\ref{tab:benchmark-composition} gives the frozen composition.

\subsection{Entities, Mentions, and Scholarly Gold}
\label{sec:benchmark-registry}

MHER is built from a source-linked registry of candidate historical persons
and attested or scholarly normalized name mentions drawn from multilingual and
multiscript sources concerning the Mongol world. Canonical identifiers and
curator-facing metadata are used only for benchmark construction and are never
exposed to evaluated systems.

A mention record separates the model-visible name representation from its
source association. This distinction between \emph{entity} and \emph{mention}
is fundamental to the benchmark: one historical person may have multiple
attestations whose forms differ substantially across languages, scripts,
transcription systems, and later scholarship.

Eighty-four persons satisfy the prespecified eligibility criteria for the
primary binary benchmark. Gold identity is inherited from the curated
scholarly registry rather than from string similarity or model behavior. Two
mentions receive a \textsc{same} label only when both are assigned to the same
confirmed historical person; a \textsc{different} label requires two distinct
confirmed persons.

For each primary entity, curator-facing evidence records link the scholarly
identification to supporting sources and uncertainty notes. These records
provide the provenance layer between historical scholarship and computational
gold labels and are not released in this arXiv version. All 84 primary entities
underwent a systematic curator re-review against the registered evidence. We
do not treat unresolved scholarly identifications as binary gold: ten such
cases are retained separately as an expert-unresolved challenge rather than
forced into either \textsc{same} or \textsc{different}. The limits of
curator-based gold validation are discussed in Section~\ref{sec:limitations}.

\subsection{Entity-Disjoint Splits and the Name-Only Core}
\label{sec:name-core}

We split the benchmark by canonical person rather than by mention pair.
Seventeen primary entities are assigned to DEV and 67 to TEST, so no
historical person occurs in both splits. This prevents different aliases or
transcriptions of the same person from appearing on opposite sides of the
development/test boundary.

Exact-surface hard-negative components are also kept within a single split.
This matters because identical or near-identical names belonging to different
people constitute an important diagnostic class: distributing the associated
entities across DEV and TEST could indirectly expose a difficult identity
contrast during development.

Evaluation pairs are generated only after the entity registry and split are
fixed. To limit overrepresentation of entities with many registered variants,
we cap the number of retained evaluation pairs for any unordered canonical
entity pair. The resulting Name-only core contains 396 balanced binary items:
80 in DEV (40 \textsc{same}, 40 \textsc{different}) and 316 in TEST
(158/158).

The 316-item TEST set serves as the benchmark's broadest evaluation of
name-form reconciliation. It includes cross-language and cross-script
variation together with deliberately confusable negatives. Challenge
properties---including cross-language, cross-script, near-surface, and
identical-surface different-person cases---are assigned before model
evaluation rather than inferred retrospectively from observed errors.

We additionally assign the 84 primary entities to coarse
\textsc{head}, \textsc{mid}, and \textsc{long-tail} prominence strata
(14, 41, and 29 entities, respectively). These categories are intended only
as a prespecified proxy for public visibility and possible model familiarity;
they are not measurements of actual pretraining frequency.

\subsection{The Source-Grounded Paired Subset}
\label{sec:source-grounded-construction}

The Source-grounded condition is not created by simply attaching arbitrary
background text to the Name-only items. It uses a stricter provenance
criterion in which historical evidence is associated with individual
\emph{mention $\times$ source} records.

For a mention $m_i$, its model-visible evidence $e_i$ is derived from the
source that attests that mention or from a separately registered source
explicitly linked to it. Evidence may describe chronology, office or role,
lineage or group affiliation, kinship, place, or participation in historical
events. Its purpose is to provide historically relevant evidence from which
identity may be inferred, not to state the identity relation itself.

Several constraints therefore govern inclusion. Model-visible evidence may
not explicitly state that two names are ``the same person,'' ``identified
with,'' or otherwise equivalent. It may not reveal registered alternative
names of the target entity beyond the surface form supplied on that side.
Contexts are not copied from a shared canonical entity description. For
\textsc{same} pairs, the two sides must additionally be independently
attested or independently sourced; if suitable evidence cannot be established
for both sides, the item remains in the Name-only core but does not enter the
paired Source-grounded subset.

These requirements yield 160 provenance-complete paired items: 52 in DEV
(26/26) and 108 in TEST (54/54). The 108 TEST items therefore form a strict
subset of the 316-item Name-only TEST set. All primary comparisons between
Name-only and Source-grounded evidence use these same 108 underlying identity
questions; the full 316-item Name-only TEST set is reported separately as a
breadth evaluation and is never used as the denominator for the paired
intervention.

The Source-grounded TEST subset was fixed without consulting model outputs.
All eligible \textsc{same} items were retained. Among
\textsc{different} candidates, prespecified identical-surface hard negatives
were retained obligatorily, while the remaining items were selected by a
frozen, model-independent procedure designed to favor contextually confusable
rather than trivially separable negatives. The selection procedure was fixed
before formal TEST evaluation and did not use TEST outcomes.

\begin{table}[t]
\centering
\caption{
Frozen MHER benchmark composition. Source-grounded, Context-only, and
shuffled-context conditions operate on the same paired items. The expert-
unresolved challenge is separate from the primary binary benchmark.
}
\label{tab:benchmark-composition}
\begin{tabular}{lrrr}
\toprule
Component & DEV & TEST & Total \\
\midrule
Primary entities
    & 17 & 67 & 84 \\
Name-only pairs
    & 80 & 316 & 396 \\
\quad \textsc{same}
    & 40 & 158 & 198 \\
\quad \textsc{different}
    & 40 & 158 & 198 \\
Source-grounded pairs
    & 52 & 108 & 160 \\
\quad \textsc{same}
    & 26 & 54 & 80 \\
\quad \textsc{different}
    & 26 & 54 & 80 \\
Context-only
    & 52 & 108 & 160 \\
Shuffled context A/B/C
    & $3\times52$ & $3\times108$ & 480 \\
Expert-unresolved
    & -- & 10 & 10 \\
\bottomrule
\end{tabular}
\end{table}

\subsection{Leakage-Aware Context Construction}
\label{sec:leakage-repair}

Source-grounded benchmarks introduce a methodological risk that does not arise
in the same form for surface-only evaluation: context construction can
accidentally encode the desired label.

A preliminary DEV construction reused entity-level context across mentions of
the same person, creating a shortcut whereby \textsc{same} pairs could be
partially identified from duplicated contextual material. We therefore
invalidated this construction and rebuilt the source-grounded condition at the
mention$\times$source level, requiring independently attested evidence for
each side. The repaired benchmark was subsequently audited against exact
duplication, explicit equivalence cues, registered self-name leakage, and
lexical or contextual shortcut signals.

This repair changes the interpretation of provenance in MHER. Provenance is
not merely bibliographic metadata attached after a benchmark item has been
created. It is a constraint on what evidence is permitted to enter the
evaluation input. The source associated with a context determines whether the
context constitutes legitimate evidence for that mention and whether two
sides of a \textsc{same} pair provide genuinely independent support.

Across the repaired Source-grounded DEV and TEST sets, the final audit found
no source-independence failure among \textsc{same} pairs, no exact context
duplication that predicts the gold relation, no explicit model-visible
equivalence statement, and no registered target-alias leakage. Prespecified
simple contextual shortcuts were also checked before formal TEST execution and
remained below the benchmark-failure criterion. Detailed audit inventories are
retained in the frozen project state and are not released in this timestamp
version.

The preliminary construction is excluded from all analyses in this paper.
Only the repaired mention$\times$source benchmark is used in DEV results,
formal TEST evaluation, controls, and downstream statistical analysis.

\subsection{Freezing, Rebuildability, and Release}

Benchmark construction was completed before formal TEST execution. Entity
splits, pair membership, source-grounded contexts, model-visible inputs,
prompts, and evaluation requests were frozen before the corresponding TEST
runs. No TEST prediction was used to alter benchmark membership, context
acquisition, balancing, challenge labels, or item selection.

A frozen project state preserves the benchmark construction, evaluation
inputs, execution records, and analysis outputs used for the reported results.
At the time of this arXiv version, the benchmark data, item-level materials,
exact execution recipes, code, and reproducibility package are not publicly
released. They are planned for public release upon publication, subject to
source-redistribution constraints.

MHER does not redistribute scans or extended copyrighted passages from modern
scholarly editions. Upon release, model-visible historical evidence will be
distributed as project-authored paraphrases accompanied by bibliographic
citations and source locators. This preserves the provenance required for
benchmark auditing while avoiding redistribution of third-party source text.

\section{Experimental Design}
\label{sec:experimental-design}

Our experiments are designed around a controlled change in the evidence
available for the same historical identity question. The primary empirical
quantity of interest is therefore not a ranking between models, but the
within-item change in reconciliation behavior when correctly grounded
historical evidence is introduced. We complement this paired intervention
with lexical and semantic baselines, context ablations, deliberately
mismatched evidence, explicit abstention, and repeated-entity robustness
analyses.

\subsection{Evaluation Conditions}
\label{sec:evaluation-conditions}

We evaluate five generative systems under a common set of evidence conditions.
The full Name-only TEST set contains 316 binary items and provides the broadest
evaluation of cross-language and cross-script name reconciliation. All
comparisons involving historical context are instead performed on the same
108-item provenance-complete TEST subset introduced in
Section~\ref{sec:source-grounded-construction}.

For a paired benchmark item with mentions $(m_a,m_b)$ and correctly associated
historical evidence $(e_a,e_b)$, we construct the following conditions.

\paragraph{Name-only.}
The system receives the two mention representations and their
language/script metadata, but no historical context:

\begin{equation}
    x_i^{\mathrm{name}}
    =
    (m_a,m_b).
\end{equation}

The full 316-item TEST set is used to characterize Name-only behavior broadly.
The corresponding 108-item subset is used whenever Name-only performance is
compared directly with a contextual condition.

\paragraph{Context-only.}
Mention surfaces and language/script metadata are removed, leaving only the
two correctly associated historical contexts:

\begin{equation}
    x_i^{\mathrm{context}}
    =
    (e_a,e_b).
\end{equation}

This ablation tests how much of the reconciliation signal is contained in the
historical evidence independently of the names themselves.

\paragraph{Source-grounded.}
The system receives both mentions together with their correctly associated
source-grounded evidence:

\begin{equation}
    x_i^{\mathrm{source}}
    =
    \big((m_a,e_a),(m_b,e_b)\big).
\end{equation}

Because the underlying identity relation is identical to the Name-only
condition, the contrast

\begin{equation}
    x_i^{\mathrm{name}}
    \longrightarrow
    x_i^{\mathrm{source}}
\end{equation}

constitutes the primary paired evidence intervention.

\paragraph{Shuffled context.}
To probe model behavior when historical text is present but pair-specific
grounding is invalidated, we construct three deterministic mismatched-context
controls:

\begin{equation}
    x_{i,r}^{\mathrm{shuffle}}
    =
    \big((m_a,\tilde e_{a,r}),
          (m_b,\tilde e_{b,r})\big),
    \qquad
    r\in\{A,B,C\}.
\end{equation}

The three derangements are deterministic and were frozen before formal TEST.
They exclude fixed-point/self-donation cases and trivial registered-alias
leakage while preserving the context pool within the split and replacing the
original mention--evidence association. The condition-specific instruction
explicitly discloses the derangement. These conditions therefore serve as
informed misgrounding stress tests rather than blinded estimates of the effect
of evidence alignment alone. The three shuffles are evaluated separately
rather than treated as independent benchmark observations.

\paragraph{Expert-signaled unresolved sanity check.}
Finally, ten expert-designated unresolved identity questions are evaluated
separately from the binary core. These items are presented in the Name-only
condition. The condition-specific instruction explicitly states that the
scholarly identity relation is not securely resolved and that a binary answer
is not required. This condition therefore serves as a protocol sanity check
for whether systems can use the available \textsc{ambiguous} action under an
explicitly signaled unresolved setting; it is not a blinded test of autonomous
uncertainty detection.

We do not construct a Source-grounded unresolved condition because the
available scholarly uncertainty notes contain explicit equivalence or
uncertainty language that would leak the intended abstention target.

Taken together, each generative model receives 316 Name-only TEST items,
five 108-item contextual conditions (Context-only, Source-grounded, and three
shuffles), and ten unresolved items.

\subsection{Lexical, Transliteration, and Embedding Baselines}
\label{sec:baselines}

We include non-generative baselines to determine how far MHER can be solved
through surface correspondence or generic semantic similarity without
generative historical reasoning. All thresholds and fitted parameters are
selected using DEV only; TEST labels are never used for model selection.

\paragraph{Name-form baselines.}
We evaluate exact surface matching, Unicode normalization and casefolding,
normalized Levenshtein similarity, Jaro--Winkler similarity, and character
$2$--$5$-gram TF--IDF similarity. Because script variation is central to the
task, we additionally apply deterministic Unidecode conversion before the
same edit- and TF--IDF-based comparisons. Unidecode is used here as a
reproducible cross-script ASCII fold, not as a historically faithful
transliteration system.

These baselines probe an increasingly permissive hierarchy of surface
correspondence: exact identity, Unicode-normalized identity, approximate
within-script similarity, and approximate cross-script similarity.

\paragraph{Context baselines.}
For the paired contextual subset, we evaluate character $2$--$5$-gram
TF--IDF, word $1$--$2$-gram TF--IDF, token Jaccard similarity, sequence
similarity, context-length features, and a standardized six-feature logistic
classifier. The latter is evaluated by fixed five-fold out-of-fold prediction
on DEV and then refit on the complete DEV set before the locked TEST run.

These methods test whether contextual reconciliation can be reproduced by
generic lexical overlap or document-level similarity rather than by combining
historically relevant evidence.

\paragraph{Multilingual semantic embedding baseline.}
We additionally evaluate the first-party Google
\texttt{gemini-embedding-2} model as a multilingual semantic-similarity
baseline under a fixed configuration. Separate decision thresholds for
Name-only and Context conditions are selected on DEV and frozen before TEST
execution. No TEST item is accessed during threshold selection.

\subsection{Generative Systems}
\label{sec:systems}

The formal generative matrix contains five systems spanning two first-party
commercial providers and one open-weight local model.

We evaluate three GPT-5.6 variants---\textbf{GPT-5.6 Terra},
\textbf{GPT-5.6 Luna}, and \textbf{GPT-5.6 Sol}---through the official
OpenAI Responses API. Returned model identifiers are recorded for every
completed response. As an independent commercial model family, we evaluate
\textbf{Gemini 3.7 Flash} through the first-party Google Gemini API.

For open-weight replication, we use \textbf{Qwen3-8B} in a frozen local
quantized deployment. Exact artifact identity, runtime, and decoding provenance
are preserved in the frozen project record but are not released in this arXiv
version.

The five systems intentionally span substantially different model sizes,
providers, access modes, and decision policies. We do not assume that their
absolute Name-only scores are directly comparable as pure measures of
historical knowledge; the primary question is whether the same evidence
manipulation produces consistent changes within each system.

\subsection{Prompting, Prediction, and Abstention}
\label{sec:prediction-protocol}

All generative systems receive the same semantic task instruction and the
same output contract, adapted only where required by provider-specific API
syntax. For each item, the prompt states that the task is to determine whether
two historical person-name attestations refer to the same individual under
the evidence provided. Except for the frozen condition-specific instruction, no gold label,
canonical entity identifier, preferred scholarly name, prominence stratum,
item-specific challenge-slice annotation, or other hidden benchmark metadata
is exposed to the model. The \textsc{shuffled-context} and
\textsc{expert-unresolved} instructions explicitly disclose their respective
control conditions. Exact prompt text is retained in the frozen project record
and is not released in this arXiv version.

Each system is instructed to return an explicitly elicited confidence
distribution over the three available actions,

\begin{equation}
    \mathbf{p}_i =
    \left(
        p_i^{\mathrm{same}},
        p_i^{\mathrm{different}},
        p_i^{\mathrm{ambiguous}}
    \right),
\end{equation}

with

\begin{equation}
    p_i^{\mathrm{same}}
    + p_i^{\mathrm{different}}
    + p_i^{\mathrm{ambiguous}}
    = 1.
\end{equation}

These values are self-reported confidence estimates generated in the required
JSON output. They are not token-level log probabilities or native likelihoods
returned by the provider APIs. We nevertheless normalize and score them as
probabilistic forecasts of the requested task actions.

A short natural-language rationale is also requested, but rationales are not
used to determine correctness. They are retained only for qualitative error
analysis.

The categorical action is defined deterministically as

\begin{equation}
    \hat y_i
    =
    \arg\max_{a\in
    \{\textsc{same},\textsc{different},\textsc{ambiguous}\}}
    p_i^a.
\end{equation}

If the maximum probability is tied exactly, the prediction is mapped to
\textsc{ambiguous}. As defined in Section~\ref{sec:abstention},
\textsc{ambiguous} is an abstention action rather than a third gold identity
relation.

Provider-specific decoding parameters and retry policies are fixed before
formal TEST execution. Successful semantic predictions are never selectively
rerun. Retries are permitted only for prespecified transport, quota,
incomplete-generation, or output-format failures. Raw provider responses are
stored before normalization so that every scored prediction can be recovered
from its original output.

\subsection{Evaluation Metrics}
\label{sec:metrics}

For the binary benchmark core, the primary outcome is ordinary accuracy.
Because the gold relation is always either \textsc{same} or
\textsc{different}, an \textsc{ambiguous} action counts as unresolved and
therefore incorrect under this metric.

We additionally report macro-F1 over the two gold identity classes. Abstaining
predictions contribute as failures to recover the corresponding gold class
rather than being evaluated as a third gold category.

To separate incorrect resolution from refusal to resolve, we report the
\emph{abstention rate}

\begin{equation}
    A
    =
    \frac{1}{n}
    \sum_{i=1}^{n}
    \mathbb{I}
    [\hat y_i=\textsc{ambiguous}],
\end{equation}

and \emph{selective accuracy}, defined as accuracy conditional on the model
making a non-abstaining \textsc{same}/\textsc{different} decision.

We additionally evaluate the quality of the explicitly elicited confidence
estimates. These quantities should be interpreted as diagnostics of
self-reported task confidence rather than as calibration of native
token-level model probabilities. For each item we record the
probability assigned to the gold relation, and summarize its mean across the
evaluation set. Multiclass Brier score and log loss are computed over the
three model actions using a one-hot target on the binary gold relation.
Probability assigned to \textsc{ambiguous} is therefore penalized when the
benchmark relation is known.

For the expert-unresolved challenge, we instead report the proportion of
items assigned \textsc{ambiguous}, the complementary over-resolution rate,
and the probability mass assigned to \textsc{ambiguous}. These quantities are
descriptive because the challenge contains only ten items.

\subsection{Paired Statistical Analysis and Repeated-Entity Robustness}
\label{sec:statistics}

The primary inferential estimand is the within-item difference between
Name-only and Source-grounded performance on the same 108 TEST questions.
The full 316-item Name-only TEST set is used only for descriptive breadth and
is never compared directly with the 108-item Source-grounded set as though
the denominators were interchangeable.

For each model, we tabulate the paired transitions

\begin{equation}
    \text{wrong}\rightarrow\text{correct},
    \qquad
    \text{correct}\rightarrow\text{wrong},
\end{equation}

together with correct$\rightarrow$correct and wrong$\rightarrow$wrong.
The primary hypothesis test is an exact two-sided McNemar test over the
discordant pairs.

Condition-specific accuracy is accompanied by Wilson 95\% confidence
intervals. The change in accuracy is summarized by a paired item-level
bootstrap with 200{,}000 resamples under a frozen resampling configuration. Because the
Name-only--Source-grounded comparison is evaluated independently for five
prespecified generative models, we also report Bonferroni-adjusted
McNemar $p$-values across the five model-level primary comparisons.

The same items may occur in multiple mention pairs through a shared canonical
entity, so pair-level observations are not the only plausible unit of
dependence. We therefore supplement the primary item-level analysis with an
entity-cluster bootstrap and repeated one-pair-per-entity sensitivity
analysis. We additionally inspect leave-one-entity-out changes to determine
whether a main effect is disproportionately driven by any individual
historical person. These analyses are robustness checks on the paired
estimand rather than replacements for it.

For the Context-only and shuffled-context analyses, all comparisons remain
paired on the same 108 items. The three shuffled controls are reported
individually; averaging across them is used only for compact descriptive
summaries, not to treat the three deterministic derangements as independent
replications.

Predefined cross-language, cross-script, near-surface, and identical-surface
different-person slices are analyzed separately. Prominence-stratified
analyses use the benchmark's preassigned \textsc{head}/\textsc{mid}/
\textsc{long-tail} labels and are treated as secondary sensitivity analyses,
not as direct evidence of training-data memorization.

\subsection{Frozen Formal Evaluation}
\label{sec:frozen-evaluation}

DEV is used for benchmark auditing, threshold selection, baseline fitting,
prompt validation, and execution checks. Formal TEST evaluation begins only
after benchmark membership, model-visible contexts, prompts, decision rules,
baseline thresholds, and analysis code are frozen.

Dedicated formal runners do not use TEST gold labels to construct requests or
select outputs. Successful model responses are not rerun based on their
semantic content, and benchmark items are never modified in response to model
performance.

The complete formal matrix contains seven conditions per generative system:
316 Name-only items, 108 Context-only items, 108 correctly Source-grounded
items, three sets of 108 shuffled-context items, and ten expert-unresolved
items, for 866 predictions per model and 4,330 formal generative predictions
overall.

All benchmark splits, contexts, prompts, and evaluation inputs were frozen
before formal TEST execution. The corresponding execution and scoring records
are preserved in the frozen project state; item-level materials and exact
execution details are withheld from this arXiv version and are planned for
public release upon publication.

\section{Main Results}
\label{sec:main-results}

We first establish how far historical identity can be recovered from surface
form and generic semantic similarity, then examine the full 316-item
Name-only TEST set, and finally turn to the primary paired intervention on the
108 items for which independently source-grounded evidence is available.
Results for Context-only, shuffled-context, and expert-unresolved conditions
are reserved for Section~\ref{sec:falsification}; challenge slices and model-
specific error behavior are analyzed in Section~\ref{sec:error-analysis}.

\subsection{Baselines Establish a Strong Non-generative Reference}
\label{sec:main-baselines}

Table~\ref{tab:baseline-hierarchy} summarizes the main frozen TEST baselines.
All thresholds were selected on DEV before TEST evaluation.

On the full 316-item Name-only TEST set, exact normalized matching reaches
48.42\% accuracy, Unicode Levenshtein similarity reaches 53.16\%, and
deterministic Unidecode conversion followed by Levenshtein similarity improves
to 59.81\%. The multilingual Gemini Embedding 2 baseline performs best among
the non-generative Name-only systems, reaching 64.87\% accuracy
(macro-F1 $=.625$).

The contextual baselines are stronger. On the 108-item paired TEST subset,
the best traditional context-similarity baseline, token Jaccard, reaches
76.85\% accuracy, while Gemini Embedding 2 reaches 86.11\%
(macro-F1 $=.861$). Thus both cross-script normalization and generic
multilingual semantic representation recover substantial signal, but neither
eliminates the remaining historical identity ambiguity.

\begin{table}[t]
\centering
\small
\setlength{\tabcolsep}{4.5pt}
\begin{tabular}{llrr}
\toprule
\textbf{Evidence} &
\textbf{Method} &
\textbf{Accuracy} &
\textbf{Macro-F1} \\
\midrule
\multirow{4}{*}{Name}
& Exact normalized match      & 48.42 & .326 \\
& Unicode Levenshtein         & 53.16 & .453 \\
& Unidecode + Levenshtein     & 59.81 & .598 \\
& Gemini Embedding 2          & \textbf{64.87} & \textbf{.625} \\
\midrule
\multirow{2}{*}{Context}
& Token Jaccard               & 76.85 & .758 \\
& Gemini Embedding 2          & \textbf{86.11} & \textbf{.861} \\
\bottomrule
\end{tabular}
\caption{
Frozen non-generative TEST baselines. Name-form methods are evaluated on the
full 316-item Name-only TEST set; context methods use the paired 108-item TEST
subset. All thresholds and fitted parameters were selected on DEV only.
}
\label{tab:baseline-hierarchy}
\end{table}

\subsection{Name-only Reconciliation Is Strongly Model-policy Dependent}
\label{sec:name-only-results}

The full 316-item Name-only TEST set reveals large differences in how
generative systems behave when historical identity must be inferred from name
forms alone. Table~\ref{tab:name-only-breadth} reports ordinary accuracy
together with abstention and selective accuracy.

GPT-5.6 Terra reaches 35.13\% ordinary accuracy and abstains on 63.61\% of
the benchmark. GPT-5.6 Luna and Sol are substantially more conservative:
their raw accuracies are only 6.01\% and 4.11\%, respectively, because they
abstain on 93.99\% and 95.89\% of the items. Yet when these systems do make
a binary decision, Luna and Sol are correct on every such decision, while
Terra reaches 96.52\% selective accuracy.

Gemini 3.7 Flash exhibits the same broad pattern: 16.77\% ordinary accuracy,
83.23\% abstention, and 100\% selective accuracy. Qwen3-8B behaves in the
opposite way. It never abstains on the full Name-only TEST set and therefore
achieves much higher raw coverage, but its accuracy is 67.09\%.

\begin{table}[t]
\centering
\small
\setlength{\tabcolsep}{5pt}
\begin{tabular}{lrrrr}
\toprule
\textbf{Model} &
\textbf{Acc.} &
\textbf{Macro-F1} &
\textbf{Abst.} &
\textbf{Sel. Acc.} \\
\midrule
GPT-5.6 Terra      & 35.13 & .495 & 63.61 & 96.52 \\
GPT-5.6 Luna       &  6.01 & .110 & 93.99 & 100.00 \\
GPT-5.6 Sol        &  4.11 & .079 & 95.89 & 100.00 \\
Gemini 3.7 Flash   & 16.77 & .270 & 83.23 & 100.00 \\
Qwen3-8B           & 67.09 & .671 &  0.00 & 67.09 \\
\bottomrule
\end{tabular}
\caption{
Name-only behavior on the full 316-item TEST set. Accuracy and abstention
should be interpreted jointly: the four API systems frequently decline to
resolve identities from names alone, whereas Qwen3-8B makes a binary decision
for every item. Selective accuracy is accuracy conditional on a non-abstaining
prediction.
}
\label{tab:name-only-breadth}
\end{table}

These results make raw Name-only accuracy difficult to interpret as a simple
ranking of historical knowledge. The systems occupy very different points on
the coverage--resolution spectrum: the API models frequently decline to
resolve uncertain names, whereas Qwen3-8B resolves aggressively. Name-only
performance therefore reflects not only name-form evidence and possible prior
familiarity, but also a model-specific decision policy toward uncertainty.

This distinction motivates the paired intervention below. The relevant
question is not whether one system has the highest Name-only accuracy, but
whether supplying independently grounded historical evidence changes the
decision on the \emph{same} identity questions.

\subsection{Source-grounded Evidence Produces Large Paired Gains}
\label{sec:paired-main-results}

Table~\ref{tab:main-paired} reports the primary comparison on the 108-item
paired TEST subset. The identity questions are identical between conditions;
only the available evidence changes from Name-only to Source-grounded.

All five systems improve under Source-grounded evidence. The four first-party
API systems show especially large changes. GPT-5.6 Terra rises from 32.41\%
to 100.00\%, a gain of 67.59 percentage points. GPT-5.6 Luna rises from
6.48\% to 99.07\% ($+92.59$ points), GPT-5.6 Sol from 4.63\% to 99.07\%
($+94.44$), and Gemini 3.7 Flash from 20.37\% to 100.00\%
($+79.63$).

The paired transition structure is strikingly asymmetric for these four
systems. Terra contains 73 wrong$\rightarrow$correct transitions and no
correct$\rightarrow$wrong transition; Luna contains 100 and 0; Sol 102 and
0; and Gemini 86 and 0. Exact McNemar tests are correspondingly extremely
small, and all four comparisons remain significant after correction across
the five prespecified model-level tests.

Qwen3-8B also improves, but the effect is smaller and less one-sided.
Accuracy rises from 75.00\% to 87.96\%, a gain of 12.96 points
(95\% paired-bootstrap CI $[2.78,23.15]$). Its transition table contains
23 wrong$\rightarrow$correct and nine correct$\rightarrow$wrong changes.
The raw exact McNemar test gives $p=.0201$; under the prespecified
five-model Bonferroni adjustment, this becomes approximately $p=.100$.
We therefore treat the Qwen gain as positive but individually weaker
inferential evidence than the four first-party comparisons.

\begin{table}[t]
\centering
\small
\setlength{\tabcolsep}{3.8pt}
\begin{tabular}{lrrrrrr}
\toprule
\textbf{Model} &
\textbf{Name} &
\textbf{Source} &
\boldmath$\Delta$ &
\textbf{95\% CI} &
\textbf{W$\rightarrow$C} &
\textbf{C$\rightarrow$W} \\
\midrule
Terra
& 32.41 & \textbf{100.00}
& +67.59 & [58.33, 75.93] & 73 & 0 \\

Luna
&  6.48 & \textbf{99.07}
& +92.59 & [87.04, 97.22] & 100 & 0 \\

Sol
&  4.63 & \textbf{99.07}
& +94.44 & [89.81, 98.15] & 102 & 0 \\

Gemini
& 20.37 & \textbf{100.00}
& +79.63 & [72.22, 87.04] & 86 & 0 \\

Qwen3-8B
& 75.00 & \textbf{87.96}
& +12.96 & [2.78, 23.15] & 23 & 9 \\
\bottomrule
\end{tabular}
\caption{
Primary paired TEST result on the same 108 historical identity questions.
Name and Source report ordinary accuracy (\%). $\Delta$ is the
Source-grounded minus Name-only accuracy difference in percentage points;
confidence intervals are paired item-bootstrap 95\% intervals.
W$\rightarrow$C and C$\rightarrow$W denote wrong-to-correct and
correct-to-wrong transitions, respectively.
}
\label{tab:main-paired}
\end{table}

\begin{figure}[t]
    \centering
    \includegraphics[width=\linewidth]{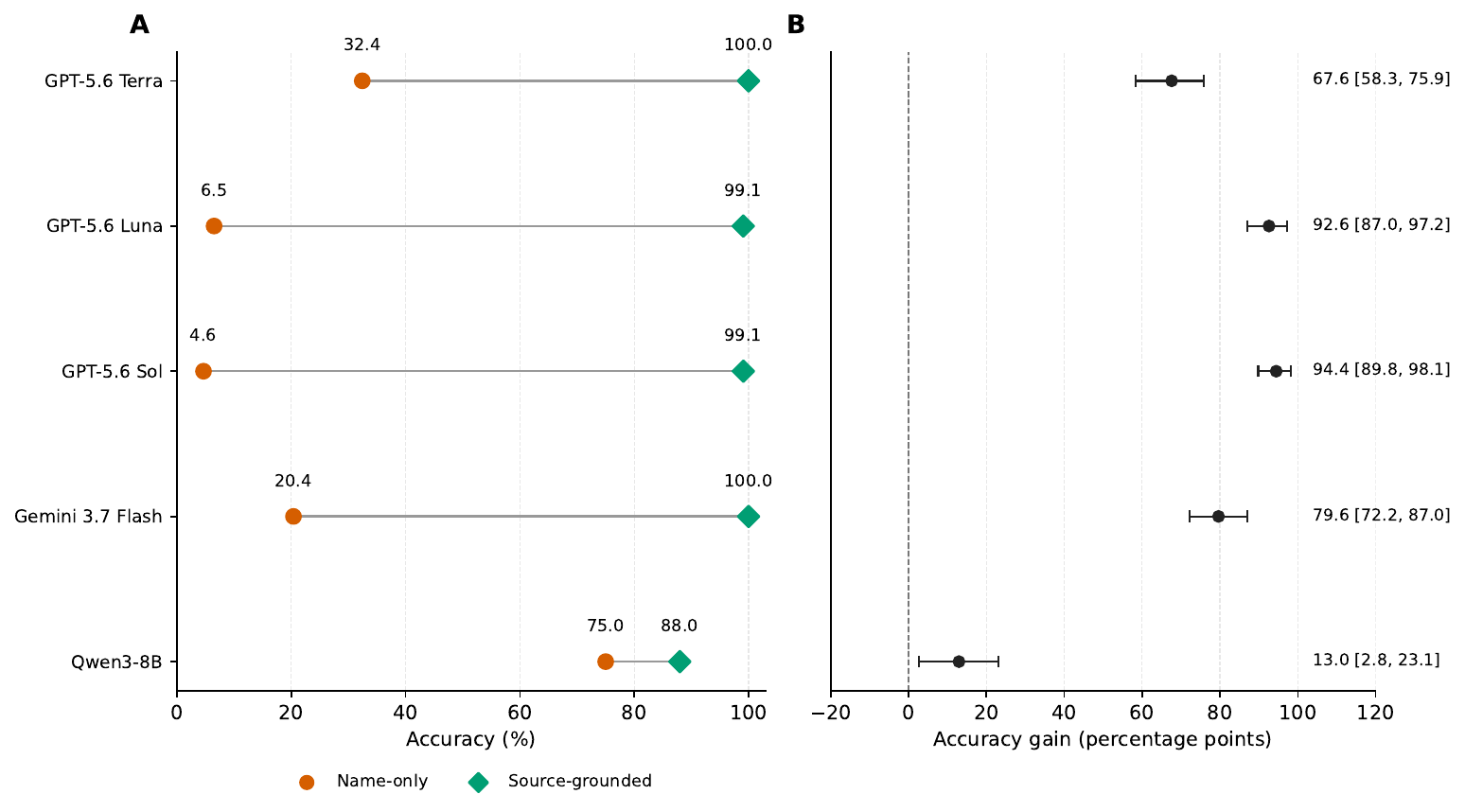}
    \caption{
Accuracy on the paired 108-item TEST subset under Name-only and
Source-grounded evidence. Panel A shows condition-specific accuracies.
Panel B shows Source-grounded minus Name-only accuracy gains with paired
item-bootstrap 95\% confidence intervals. The four first-party API systems
move from highly abstention-dominated Name-only behavior to near-ceiling
Source-grounded resolution; Qwen3-8B shows a smaller but positive paired gain.
}
    \label{fig:paired-main-results}
\end{figure}

Figure~\ref{fig:paired-main-results} visualizes the same paired contrast.
The principal pattern is not a common absolute Name-only starting point:
the five systems begin from radically different decision policies. Rather,
the common pattern is a strong movement toward correct resolution once the
identity question is accompanied by source-linked historical evidence.

The magnitude of the effect is also notable relative to non-generative
baselines. On the same contextual TEST subset, Gemini Embedding 2 reaches
86.11\% when comparing the historical contexts semantically. Source-grounded
generative performance ranges from 87.96\% for Qwen3-8B to 100\% for Terra
and Gemini. The result is therefore not explained simply by replacing
cross-script name similarity with a strong multilingual embedding score.

At the same time, the near-ceiling results of the first-party systems should
not be interpreted as evidence that historical entity reconciliation is
generally solved. The paired subset is deliberately restricted to cases for
which independently sourced and provenance-traceable evidence exists on both
sides. What the intervention demonstrates is that, within this evidence-
eligible subset, exposing the correct historical evidence can resolve
identity questions that remain highly uncertain under names alone.

\subsection{Correct Evidence Compresses Cross-model Behavioral Differences}
\label{sec:cross-model-convergence}

The five systems disagree sharply in the Name-only condition because they
adopt very different uncertainty policies. On the paired subset, Name-only
accuracy ranges from 4.63\% for Sol to 75.00\% for Qwen3-8B. Once
Source-grounded evidence is supplied, the range contracts to
87.96--100.00\%.

This convergence is important because it occurs despite substantial
differences in model family, provider, access mode, and propensity to abstain.
The result therefore does not suggest that all models behave similarly in the
absence of evidence. Instead, it indicates that correctly associated
historical evidence acts as a stabilizing signal across otherwise divergent
decision policies.


Section~\ref{sec:falsification} examines whether this convergence persists when names are removed
and when pair-specific grounding is deliberately invalidated and explicitly
disclosed. These conditions provide ablations and informed stress tests rather
than a blinded causal estimate of evidence alignment alone.
We also show there that removing the names
altogether does not produce a uniform pattern across model families, which
becomes important for interpreting the open-weight model's behavior.

\section{Falsification and Ablation Analysis}
\label{sec:falsification}

The primary intervention in Section~\ref{sec:main-results} establishes that
Source-grounded evidence changes reconciliation behavior on the same
historical identity questions. That comparison alone, however, does not show
which component of the contextual condition is responsible for the effect.
Historical context may carry sufficient identity information even without the
name forms, and arbitrary additional text could in principle introduce
dataset- or topic-level shortcuts.

We therefore analyze three complementary controls: Context-only removes name
surfaces while preserving the correctly associated historical evidence;
three deterministic shuffled-context conditions preserve the context pool
while destroying mention--evidence alignment; and an expert-signaled unresolved sanity check verifies whether the common
prediction protocol supports abstention when unresolved status is explicitly
disclosed.
Finally, we test whether the primary paired result is robust to repeated
canonical entities.

\subsection{Historical Context Carries Substantial Signal Without Names}
\label{sec:context-only-results}

The Context-only condition removes both mention surfaces and
language/script metadata while preserving the two correctly associated
historical contexts. Performance remains high across all five systems
(Table~\ref{tab:control-summary}), ranging from 80.56\% for GPT-5.6 Luna to
94.44\% for both Gemini 3.7 Flash and Qwen3-8B.

For the four first-party API systems, restoring the name forms produces a
further numerical increase: Source-grounded exceeds Context-only by
12.96 percentage points for Terra, 18.52 for Luna, 6.48 for Sol, and
5.56 for Gemini. These differences indicate that name-form evidence can
remain useful once strong historical context is available, although the
majority of the recoverable signal is already present in the historical
evidence itself.

Qwen3-8B is the important exception. Its Context-only accuracy is 94.44\%,
whereas Source-grounded accuracy is 87.96\%, a numerical decrease of
6.48 points. The paired Source-grounded--Context-only McNemar comparison is
not statistically significant ($p=.0923$), so we do not interpret this
difference as evidence of a general degradation effect. It nevertheless
shows that name surfaces are not uniformly additive across model families.
We return to the directional structure of Qwen's errors in
Section~\ref{sec:error-analysis}.

\begin{table}[t]
\centering
\small
\setlength{\tabcolsep}{3.6pt}
\begin{tabular}{lrrrrr}
\toprule
\textbf{Model} &
\textbf{Ctx.-only} &
\textbf{Source} &
\textbf{Shuffle range} &
\boldmath$\Delta_{\mathrm{S-C}}$ &
\boldmath$\Delta_{\mathrm{S-Shuf}}$ \\
\midrule
Terra
& 87.04 & 100.00 & 13.89--16.67 & +12.96 & +85.19 \\
Luna
& 80.56 & 99.07 & 11.11--22.22 & +18.52 & +82.41 \\
Sol
& 92.59 & 99.07 & 0.93--4.63 & +6.48 & +96.30 \\
Gemini
& 94.44 & 100.00 & 49.07--50.00 & +5.56 & +50.31 \\
Qwen3-8B
& 94.44 & 87.96 & 57.41--63.89 & -6.48 & +28.09 \\
\bottomrule
\end{tabular}
\caption{
Falsification and ablation results on the paired 108-item TEST set.
$\Delta_{\mathrm{S-C}}$ is Source-grounded minus Context-only accuracy.
$\Delta_{\mathrm{S-Shuf}}$ is Source-grounded minus the mean of the three
deterministic shuffled-context accuracies. The three derangements are
analyzed separately rather than treated as independent replications.
}
\label{tab:control-summary}
\end{table}

These results refine the interpretation of the primary intervention.
Source-grounded reconciliation should not be understood simply as a setting
in which names are augmented with useful background text. Historical context
itself carries substantial identity information, and the marginal value of
restoring name surfaces depends on the evaluated system.

\subsection{Correctly Grounded and Explicitly Misgrounded Evidence
Produce Divergent Behavior}
\label{sec:shuffle-results}

The shuffled-context controls provide an informed misgrounding stress test.
They preserve the same pool of historical contexts, together with its source
mix, writing style, topic distribution, and length distribution, while
replacing the original mention--evidence association with a frozen
derangement.

Because the condition-specific prompt explicitly informs models that the
contexts were drawn from other historical entities under this derangement,
the comparison is not a blinded test of whether models can autonomously
detect incorrect provenance, nor does it isolate evidence alignment from the
control instruction itself. Instead, it tests how systems behave when the
same historical context pool is presented under deliberately invalidated and
explicitly disclosed grounding.

If performance remained high even under this disclosed misgrounding
condition, that would indicate that generic historical prose or corpus-level
contextual regularities were sufficient to sustain reconciliation despite
the absence of valid pair-specific evidence.

It does not.

Across all five systems, each correctly Source-grounded endpoint lies well
above all three shuffled controls. Terra scores 100.00\% with correct
grounding but only 13.89--16.67\% after shuffling. Luna falls from 99.07\%
to 11.11--22.22\%, and Sol from 99.07\% to 0.93--4.63\%. Gemini behaves
differently in absolute terms, but the same falsification holds:
100.00\% with correct evidence versus approximately chance accuracy
(49.07--50.00\%) under the three shuffles. Qwen3-8B, which is generally more
willing to make binary decisions, falls from 87.96\% to 57.41--63.89\%.

Averaged over the three deterministic shuffles, the gap between correct
Source-grounded evidence and mismatched evidence is 85.19 percentage points
for Terra, 82.41 for Luna, 96.30 for Sol, 50.31 for Gemini, and 28.09 for
Qwen3-8B. Even for Qwen---the system with the smallest primary
Name-only--Source-grounded effect---all three exact Source-grounded versus
Shuffle comparisons strongly favor correctly grounded evidence
(raw McNemar $p \leq 4.10\times10^{-4}$).

\begin{figure}[t]
    \centering
    \includegraphics[width=\linewidth]
        {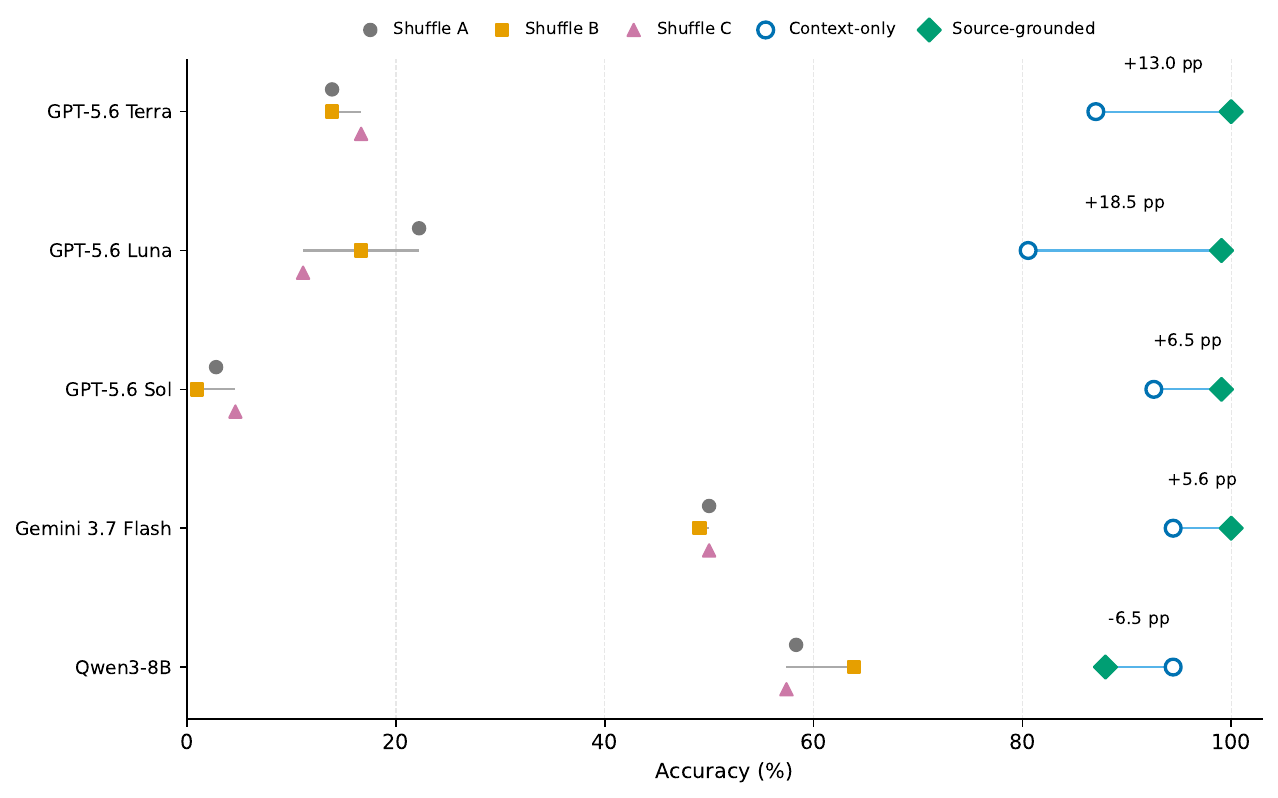}
    \caption{
    Falsification and ablation analysis on the paired 108-item TEST set.
    Horizontal ranges summarize the three deterministic shuffled-context
    controls; Context-only and correctly Source-grounded accuracy are shown
    separately. Correctly Source-grounded evidence substantially outperforms the three
explicitly signaled misgrounding controls across all five systems. The annotation beside each
    model gives Source-grounded minus Context-only accuracy.
    }
    \label{fig:falsification-ablation}
\end{figure}

The control shows that the high Source-grounded endpoint is not reproduced
when pair-specific grounding is explicitly invalidated and disclosed.
Because the Source-grounded and shuffled conditions also differ in their
condition-specific instructions, the resulting gap should not be interpreted
as a blinded causal estimate of alignment alone. It nevertheless rules out
the stronger possibility that the same historical prose is sufficient to
sustain near-ceiling performance even when it is explicitly presented as
misgrounded evidence. The same historical text can be highly informative
when associated with the correct attestations and actively unhelpful when
associated with the wrong ones. Source-grounded performance therefore cannot
be attributed simply to longer inputs, historical register, or generic
semantic richness.

The shuffled controls are therefore most informative when interpreted
together with the Context-only and Source-grounded conditions.
Context-only establishes that correctly associated historical descriptions
carry substantial identity signal; Source-grounded measures behavior when
those descriptions are paired with the corresponding names; and the shuffled
conditions characterize behavior when pair-specific grounding is deliberately
invalidated and disclosed. Their performance is not reproduced by a benchmark-wide
contextual shortcut that survives reassignment: once the mention--source
relation is broken, accuracy collapses despite preserving the underlying
context corpus. Model-specific behavior under mismatched evidence is analyzed
further in Section~\ref{sec:error-analysis}.

\subsection{Expert-Signaled Unresolved Cases Provide a Protocol Sanity Check}
\label{sec:unresolved-results}

The primary binary benchmark contains only identity relations that were
treated as sufficiently established for \textsc{same} or \textsc{different}
gold assignment. MHER-Unresolved instead contains ten identity questions
retained outside this core because the scholarly relation was not adjudicated
as a resolved binary fact.

The condition-specific instruction explicitly identifies these items as
expert-unresolved and states that a binary answer is not required. The set
therefore functions as a protocol sanity check for whether evaluated systems
can use the available \textsc{ambiguous} action when unresolved status is
explicitly signaled, rather than as a blinded test of whether models can
discover historiographical uncertainty autonomously.

All five generative systems return \textsc{ambiguous} on all ten cases:
50 abstentions in 50 model--item decisions. Thus the common prediction
protocol does not force a binary resolution when applied to this separate
expert-unresolved set.

We interpret this result narrowly. It demonstrates that the common output
protocol supports abstention as an operationally meaningful action when an
expert-designated unresolved condition is explicitly disclosed. It does not
show that the models independently inferred scholarly uncertainty from the
name forms, nor does it establish how they would behave if unresolved status
had to be diagnosed without a condition cue.

We do not construct a Source-grounded unresolved condition because the
available scholarly uncertainty notes contain explicit uncertainty or
equivalence language that would leak the desired abstention target. A stronger
future design would supply independently sourced, neutral evidence without an
explicit unresolved designation and test whether systems can identify
insufficient evidence autonomously.

\subsection{The Main Effect Is Not Driven by Repeated Entities}
\label{sec:repeated-entity-results}

Although DEV and TEST are entity-disjoint, multiple TEST pairs can still
involve the same canonical person within a split. The paired 108-item TEST
subset contains 31 canonical entities that occur in more than one item.
We therefore test whether the Name-only--Source-grounded gain is an artifact
of a small number of repeatedly represented entities.

The result is highly stable for the four first-party API systems.
Entity-cluster bootstrap mean gains are 70.37 points for Terra
(95\% CI $[65.43,75.64]$), 93.18 for Luna
($[90.48,96.30]$), 93.87 for Sol
($[92.11,96.05]$), and 77.01 for Gemini
($[73.33,80.77]$). Repeated one-pair-per-entity thinning produces similarly
large estimates, and leave-one-entity-out analysis changes the full paired
gain by at most 2.19 percentage points for any of these systems.

Qwen again shows a smaller but directionally consistent effect. Its
entity-cluster bootstrap gain is 12.08 points
(95\% CI $[7.04,16.67]$), close to the full pair-level estimate of
12.96 points. The repeated one-pair-per-entity sensitivity has a mean gain
of 13.61 points but a wider 95\% interval
($[-3.23,30.00]$), reflecting the weaker and less one-sided Qwen result.
Its leave-one-entity-out gain remains positive for every removed entity,
ranging from 9.80 to 15.53 points.

\begin{table}[t]
\centering
\small
\setlength{\tabcolsep}{3.6pt}
\begin{tabular}{lrrr}
\toprule
\textbf{Model} &
\textbf{Pair gain} &
\textbf{Cluster boot.} &
\textbf{LOO range} \\
&
\textbf{(pp)} &
\textbf{mean [95\% CI]} &
\textbf{(pp)} \\
\midrule
Terra
& +67.59 & +70.37 [65.43, 75.64] & 65.69--69.70 \\
Luna
& +92.59 & +93.18 [90.48, 96.30] & 91.92--94.17 \\
Sol
& +94.44 & +93.87 [92.11, 96.05] & 93.94--95.28 \\
Gemini
& +79.63 & +77.01 [73.33, 80.77] & 78.43--81.82 \\
Qwen3-8B
& +12.96 & +12.08 [7.04, 16.67] & 9.80--15.53 \\
\bottomrule
\end{tabular}
\caption{
Repeated-entity robustness for the primary Name-only to Source-grounded
intervention. Cluster bootstrap resamples canonical entities rather than
individual pairs. LOO reports the range of paired gains after removing each
entity in turn. Additional entity-level sensitivity outputs are retained in
the frozen project record.
}
\label{tab:repeated-entity-summary}
\end{table}

Taken together, the ablations and falsification controls support a more
specific interpretation of the main result. Historical context is itself a strong source of identity evidence, while the
near-ceiling Source-grounded endpoint is not reproduced under the explicitly
signaled misgrounding conditions. The primary effect also
persists after accounting for repeated entities and is therefore not
explained by a small number of heavily represented persons.

At the same time, the controls reveal meaningful model heterogeneity:
Source-grounded evidence is not uniformly superior to Context-only evidence,
and systems respond differently when evidence is deliberately misaligned.
Section~\ref{sec:error-analysis} examines these differences at the level of
surface-confusable cases, abstention and probabilistic behavior, and
model-specific errors.

\section{Model Behavior and Error Analysis}
\label{sec:error-analysis}

Aggregate accuracy conceals substantial differences in how the evaluated
systems use names, historical evidence, and abstention. We therefore examine
four aspects of model behavior: performance on cases in which surface-form
similarity is actively misleading; the directional error pattern of the
open-weight Qwen3-8B model; changes in abstention and probabilistic quality
across evidence conditions; and sensitivity to entity prominence.

These analyses reinforce the distinction introduced in
Section~\ref{sec:name-vs-identity}: name correspondence is evidence about
historical identity, but it is not itself the identity relation.

\subsection{Surface-confusable Cases Expose the Limit of Name Matching}
\label{sec:hard-surface}

The clearest diagnostic cases are those for which surface similarity and
historical identity point in different directions.

On the seven prespecified near-surface items in the paired TEST set, every
generative model scores 0/7 under Name-only evidence. With correctly
Source-grounded evidence, Terra, Luna, Gemini, and Qwen3-8B resolve all
seven correctly; Sol resolves six and abstains on the remaining item.

The five identical-surface \textsc{different}-person items provide an even
stronger test. Here the two visible names are exactly identical after the
benchmark's normalization, so no decision rule based on name identity alone
can distinguish the people. All five models score 0/5 under Name-only
evidence. Under Source-grounded evidence, Terra, Luna, Gemini, and Qwen3-8B
resolve all five correctly. Sol resolves four correctly and abstains on the
fifth.

Aggregated across models, the identical-surface challenge therefore changes
from

\begin{equation}
    0/25
    \quad\longrightarrow\quad
    24/25
\end{equation}

correct binary resolutions. The remaining decision is an abstention rather
than a false \textsc{same} match.

This result is important because it isolates the conceptual difference
between name matching and identity reconciliation. When two distinct
historical people have the same visible name, additional normalization,
transliteration, or string similarity cannot in principle recover the
distinction. The successful Source-grounded decisions instead depend on
evidence such as chronology, office, kinship, political role, and historical
setting.

The broader challenge slices show the same direction. Source-grounded
accuracy reaches 87.76--100\% across the 98 cross-language items and
86.67--100\% across the 60 cross-script items, despite much larger
cross-model variation under names alone. 

A representative identical-surface item contains two attestations with the
same visible name but historically incompatible chronology, kinship, and
political setting. The string itself provides no basis for separating the two
referents; the distinction becomes available only through independently
source-grounded evidence. This is precisely the class of case that MHER is
designed to evaluate.

\subsection{Qwen3-8B Exhibits Surface-form Interference and Over-resolution}
\label{sec:qwen-interference}

Qwen3-8B provides the most informative departure from the near-ceiling
behavior of the four first-party API systems. As shown in
Section~\ref{sec:context-only-results}, Qwen reaches 94.44\% accuracy under
Context-only evidence but 87.96\% when the corresponding name forms are
restored in the Source-grounded condition. The difference is not significant
under the paired McNemar test ($p=.0923$), but its error structure is highly
directional.

All 13 Qwen Source-grounded errors have gold label
\textsc{different} and are predicted \textsc{same}. Qwen nevertheless
resolves all 54 Source-grounded \textsc{same} items correctly. Its residual
failure mode is therefore not missed positive identity, but
\emph{over-resolution}: distinct historical people are incorrectly merged.

More importantly, ten of the 13 errors are cases that Qwen had resolved
correctly as \textsc{different} under Context-only evidence. Restoring the
name surfaces changes these ten decisions from correct
\textsc{different} to erroneous \textsc{same}. We refer to this observable
transition as \emph{surface-form interference}:

\begin{equation}
    \underbrace{
    \hat y_i^{\mathrm{context}}
    = \textsc{different}
    }_{\text{correct}}
    \quad\longrightarrow\quad
    \underbrace{
    \hat y_i^{\mathrm{source}}
    = \textsc{same}
    }_{\text{incorrect}}.
\end{equation}

The remaining three Source-grounded errors are persistent conflations:
Context-only already predicts \textsc{same}, and adding the name does not
repair the mistake.

We manually coded the 13 errors into three descriptive categories
(Table~\ref{tab:qwen-error-taxonomy}). These categories characterize the
observed outputs; they should not be interpreted as uniquely identified
internal causal mechanisms.

\begin{table}[t]
\centering
\small
\setlength{\tabcolsep}{4.2pt}
\begin{tabular}{lrrr}
\toprule
\textbf{Error category} &
\textbf{Name-induced} &
\textbf{Persistent} &
\textbf{Total} \\
\midrule
Unsupported alias / transliteration
    & 5 & 1 & 6 \\
Decision--rationale inconsistency
    & 4 & 0 & 4 \\
Coarse role / chronology conflation
    & 1 & 2 & 3 \\
\midrule
Total
    & 10 & 3 & 13 \\
\bottomrule
\end{tabular}
\caption{
Qualitative taxonomy of the 13 Qwen3-8B Source-grounded errors.
``Name-induced'' denotes cases that are correct under Context-only evidence
but become false \textsc{same} merges when the name surfaces are restored.
``Persistent'' denotes errors already present under Context-only evidence.
Only aggregate coding is reported in this arXiv version.
}
\label{tab:qwen-error-taxonomy}
\end{table}

The largest category consists of unsupported alias or transliteration
equivalences: after name surfaces are restored, some outputs assert an
identity relation despite contextual evidence distinguishing the two people.
A second pattern is decision--rationale inconsistency, in which the generated
explanation recognizes conflicting roles, chronology, place, or lineage while
the returned action still favors \textsc{same}. Because rationales are not
used for scoring, this does not alter the evaluation protocol; analytically,
it shows that the categorical decision can conflict with evidence recognized
in the model's own explanation. Item-level cases are retained in the frozen
project record and are not released in this arXiv version.

The Qwen result therefore adds an important qualification to a simple
``names plus context'' account. Historical context is sufficiently
informative for this model to reach 94.44\% without seeing the names, but
restoring the name surfaces can activate unsupported identity hypotheses and
produce false merges. Names are consequently not a uniformly beneficial
feature once strong historical evidence is already available.

\subsection{Evidence Changes Decision Policy and Probabilistic Behavior}
\label{sec:behavior-calibration}

\begin{figure}[t]
    \centering
    \includegraphics[width=\linewidth]{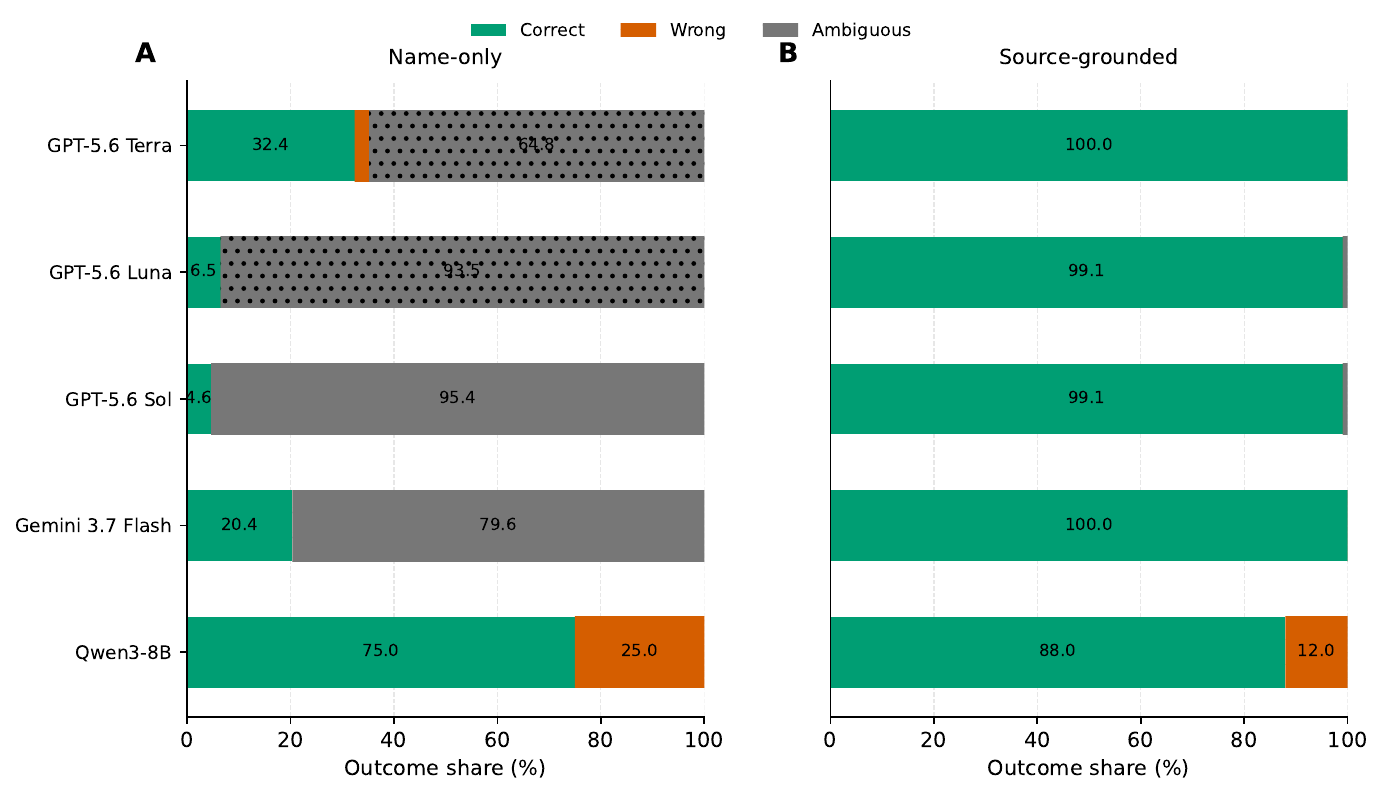}
    \caption{
    Outcome composition on the paired 108-item TEST subset under
    Name-only (left) and Source-grounded evidence (right).
    The first-party API systems are strongly abstention-dominated when names
    are presented without historical evidence, whereas Qwen3-8B makes a
    binary decision on every Name-only item. Correctly grounded evidence
    largely removes abstention for the API systems and sharply compresses
    cross-model behavioral differences.
    }
    \label{fig:model-behavior}
\end{figure}

The systems also differ substantially in \emph{how} they express uncertainty.
Figure~\ref{fig:model-behavior} decomposes the paired 108-item results into
correct, incorrect, and \textsc{ambiguous} outcomes.

Under Name-only evidence, Terra abstains on 64.81\% of the paired items,
Luna on 93.52\%, Sol on 95.37\%, and Gemini on 79.63\%. Qwen3-8B abstains
on none. Thus superficially low Name-only accuracy for the first-party
systems is primarily a coverage phenomenon, whereas Qwen's errors reflect
aggressive binary resolution.

Correct Source-grounded evidence changes this profile almost completely.
Terra and Gemini abstain on no paired item; Luna and Sol each abstain on only
one of 108. Their remaining decisions are otherwise correct. Qwen continues
to make no abstaining predictions, but retains the 13 false
\textsc{same} merges discussed above.

Metrics computed from the explicitly elicited confidence estimates show the same broad stabilization for the
first-party systems. Under Source-grounded evidence, their mean probability
assigned to the gold relation ranges from .966 to .986, with Brier scores
between .001 and .012. Qwen assigns a lower mean gold probability of .792
and has a Brier score of .202.

The Qwen Context-only/Source-grounded contrast is particularly informative.
Its mean $p(\mathrm{gold})$ actually rises from .732 to .792 when names are
restored, yet its Brier score worsens from .164 to .202 and log loss from
.352 to .385. Mean gold probability alone therefore obscures a concentrated
tail of relatively confident false merges; proper scoring rules expose this
loss of probabilistic quality.

Mismatched evidence produces yet another form of model heterogeneity. Under
the three shuffled controls, the OpenAI systems respond predominantly by
abstaining: Terra abstains on 80.56--85.19\% of shuffled items, Luna on
77.78--87.96\%, and Sol on 92.59--96.30\%. Gemini instead abstains on only
2.78--6.48\% and remains near chance in ordinary accuracy, while exhibiting
very poor probabilistic scores (Brier approximately .845--.853; log loss
3.07--4.25). Qwen never abstains under shuffling and retains 57.41--63.89\%
accuracy, but its Brier scores degrade to approximately .571--.647.

Under the explicitly signaled misgrounding conditions, there is no universal
failure response.
Depending on the model, mismatched evidence can induce abstention, continued
binary resolution with poor probabilistic quality, or false merging. What is shared across systems is not the failure mode but the contrast between
correctly Source-grounded and explicitly signaled misgrounded historical
evidence.

\subsection{Prominence Sensitivity Is Model-specific}
\label{sec:prominence-results}

A final question is whether Name-only reconciliation is disproportionately
successful for historically prominent people. Because a
\textsc{different} pair involves two entities that may belong to different
prominence strata, we evaluate this question on the 158
\textsc{same} items in the full Name-only TEST set and first average
accuracy within canonical entity. Sixty-six TEST entities contribute to this
analysis.

The relationship is not universal across models. Entity-macro Spearman
correlations between ordinal prominence and Name-only accuracy are small for
Terra ($\rho=.091$, $p=.466$) and Sol ($\rho=.073$, $p=.561$).
Luna exhibits the strongest positive association
($\rho=.350$, $p=.0039$), which remains significant after Bonferroni
correction across the five models ($p_{\mathrm{adj}}=.0197$).
Gemini ($\rho=.297$, $p=.0154$) and Qwen3-8B
($\rho=.247$, $p=.0458$) show nominal positive associations, but neither
survives the same five-model correction.

These patterns support the narrower conclusion that
\emph{Name-only prominence sensitivity is model-specific}. They do not
establish that any particular effect is caused by pretraining memorization.
The prominence labels are coarse proxies for public visibility rather than
measurements of actual training exposure, and Name-only accuracy also mixes
surface correspondence with model-specific abstention policies.

Within the 54 \textsc{same} items in the Source-grounded paired subset,
all five models resolve every item correctly across the
\textsc{head}, \textsc{mid}, and \textsc{long-tail} strata. This shows that
the available evidence is sufficient to support successful same-person
reconciliation throughout the three prominence groups in this selected
subset. The ceiling, however, prevents meaningful estimation of a residual
prominence gradient, and the source-eligible subset should not be interpreted
as representative of all historically documented entities.

\subsection{Summary}
\label{sec:behavior-summary}

The error analyses reveal a consistent distinction between surface evidence
and historically grounded identity evidence.

First, the hardest surface-form cases are not merely quantitatively difficult:
identical names belonging to different people are structurally impossible to
resolve from name identity alone, yet 24/25 model--item decisions become
correct under Source-grounded evidence, with the remaining decision an
abstention.

Second, the Qwen3-8B errors show that name information can sometimes
\emph{interfere} with otherwise sufficient historical evidence. Ten of its
13 Source-grounded errors arise specifically when restoring names converts a
correct Context-only distinction into a false identity merge.

Third, uncertainty behavior is strongly model-dependent. Some systems respond
to insufficient or mismatched evidence by abstaining, while others continue
to make binary decisions. Correctly grounded evidence nevertheless produces
a marked convergence in both accuracy and probabilistic quality.

Finally, prominence effects under names alone vary across systems rather than
forming a universal familiarity gradient. Together, these findings suggest
that historical entity reconciliation should be analyzed not only in terms
of whether a model is correct, but also in terms of which evidence it uses,
when it chooses to resolve, and how surface-form priors interact with
source-grounded historical evidence.

\section{Discussion}
\label{sec:discussion}

MHER was designed around a simple distinction: historical names are evidence
about identity, but they are not themselves the identity relation. The
experiments make this distinction observable. Systems that behave very
differently when only name forms are visible become substantially more
consistent when given correctly associated historical evidence. Conversely,
when the same context pool is reassigned under an explicitly signaled
misgrounding control, performance decreases sharply across model families.
At the same time, the Context-only and Qwen3-8B results show that names are
not uniformly necessary or uniformly beneficial once informative historical
evidence is available.

We discuss what these findings imply for historical entity processing,
evidence-grounded evaluation, benchmark design, and the interpretation of
contemporary language models in historical research.

\subsection{From Name Familiarity to Evidence-Conditioned Reconciliation}
\label{sec:discussion-evidence}

A common way to evaluate cross-lingual or cross-script entity systems is to
ask whether the system can recover a canonical identity from a surface form.
This framing is natural when a stable target inventory exists and when the
principal challenge lies in transliteration, lexical variation, or candidate
retrieval. Historical identity, however, often presents a different problem:
the surface form may be ambiguous, the relevant individual may be absent from
a modern knowledge base, and the evidence needed for identification may be
distributed across chronology, kinship, office, geography, and event
participation.

The MHER results suggest that these two settings should be interpreted
differently. Name-only performance mixes several sources of signal:
orthographic correspondence, learned transliteration regularities, possible
prior familiarity with the individual, and a system-specific willingness to
commit to an identity judgment. The extreme variation in abstention across
models demonstrates that a Name-only score is therefore not a pure measure of
historical knowledge.

Source-grounded evaluation changes the question. Instead of asking whether a
model already ``knows'' that two surface forms belong to the same historical
person, it asks whether the model can use supplied historical evidence to
support or reject an identity relation. This is closer to the logic of
historical record linkage, where noisy or non-unique names are combined with
additional biographical attributes rather than treated as sufficient
identifiers in isolation \citep{fellegi1969theory,abramitzky2020automated}.

This distinction is especially visible in the identical-surface hard
negatives. When two distinct individuals share the same visible name, better
normalization or transliteration cannot recover a distinction that is absent
from the string itself. Historical evidence can. In this sense, the relevant
computational capability is not merely robust name matching but
\emph{evidence-conditioned identity reconstruction}.

The strong Context-only results sharpen this interpretation further.
Historical evidence often carries most of the information required for
reconciliation even after the names themselves are removed. The role of the
name is therefore conditional rather than foundational: it may provide useful
additional evidence, but it does not define the identity relation. Qwen3-8B's
surface-form interference provides the converse case, showing that a name can
sometimes introduce a misleading prior into an otherwise sufficient
contextual judgment.

\subsection{Implications for Historical NLP and Digital Prosopography}
\label{sec:discussion-history}

The findings have implications beyond the specific Mongol-world benchmark.
Historical NLP frequently operates in settings where documentary evidence is
fragmentary, orthography is unstable, and modern entity inventories provide
uneven coverage. Recent historical entity-linking resources have begun to
address long-tail knowledge, non-Latin historical text, literary entities, and
multilingual historical linking
\citep{blouin2024dataset,graciotti2025kemhisto,sarkar2025mahanama,santini2026confidence}.
MHER complements this line of work by focusing on an identity relation between
two source attestations rather than requiring either mention to resolve first
to a canonical KB target.

This pairwise view is closely aligned with long-standing practices in
prosopography and historical record linkage. Historical scholars rarely
identify people from names alone. Identity is reconstructed from combinations
of family relations, offices, locations, dates, institutional affiliations,
and co-occurring actors. Large historical linkage projects likewise show that
supplementary family and geographic information can materially improve person
matching when names and demographic fields are ambiguous
\citep{abramitzky2020automated}.

For computational historical research, this suggests a practical shift in
what should count as useful model capability. A system that recognizes famous
rulers from their names may be useful, but historical corpora are also
populated by officials, relatives, envoys, regional actors, scribes, and other
long-tail individuals whose identities may not be strongly represented in
pretraining data. For such cases, the more transferable capability may be the
ability to combine explicit source evidence rather than to rely on stored name
associations.

Our prominence results are consistent with this interpretation but should not
be read as direct evidence about memorization. Name-only prominence
sensitivity varies by model, and the prominence labels are only proxies for
public visibility. What the Source-grounded results show more directly is that
the selected historical evidence can support successful reconciliation across
prominence strata even when Name-only behavior differs substantially between
systems.

This point matters for digital humanities applications. If computational
systems are to assist with archival aggregation, prosopographic database
construction, or cross-source person reconciliation, a desirable workflow is
not one in which the model silently substitutes its parametric memory for the
archive. A more auditable workflow exposes the evidence used for the
identification and preserves the provenance of that evidence so that a
historian can inspect, contest, or revise the reconciliation.

\subsection{Grounding Is Part of the Benchmark Semantics}
\label{sec:discussion-grounding}

The contextual controls highlight a broader methodological point:
\emph{providing context is not equivalent to providing valid evidence}.

This distinction is familiar from retrieval-augmented and attributed language
modeling. External documents can improve knowledge-intensive generation
\citep{lewis2020retrieval}, but the presence of retrieved material does not by
itself establish that a prediction is grounded in the correct evidence. Work
on attributed and citation-supported generation has consequently emphasized
the relationship between a model output and its supporting sources
\citep{gao2023enabling,malaviya2024expertqa}.

MHER exposes a related distinction at the benchmark-input level. Context-only
performance shows that historical descriptions can be highly informative.
The shuffled conditions preserve the historical context pool while replacing
the original mention--evidence association with a frozen derangement. Under
these conditions, performance decreases sharply across model families.

Importantly, this comparison is an \emph{informed misgrounding stress test},
not a blinded test of whether models can autonomously discover incorrect
provenance. The condition-specific prompt explicitly informs the evaluated
system that the supplied contexts were drawn from other historical entities
under a frozen derangement. The result therefore demonstrates that systems
behave very differently when evaluated with correctly aligned evidence versus
explicitly identified misgrounded evidence; it does not isolate the effect of
evidence alignment independently of the condition instruction.

The relevant conceptual distinction remains between

\begin{equation}
    \text{historical context being present}
\end{equation}

and

\begin{equation}
    \text{historical context constituting valid evidence for the mention}.
\end{equation}

This has consequences for benchmark construction. If contextual evidence is
part of an evaluation, provenance cannot be treated as optional documentation
added after dataset creation. It determines the semantics of the input. A
context associated with the wrong person may be fluent, historically
plausible, and topically relevant while nevertheless constituting invalid
evidence for the identity question being asked.

The preliminary leakage-prone MHER construction illustrates the complementary
risk. Reusing entity-level contextual material made shared context itself
predictive of \textsc{same}. The subsequent mention$\times$source redesign
therefore reflects a general principle: contextual benchmarks require controls
not only against label leakage in the text, but also against leakage induced
by how evidence is assigned to examples.

This connects MHER to a broader literature on annotation artifacts, heuristic
shortcuts, and controlled challenge sets
\citep{gururangan2018annotation,mccoy2019right,gardner2020evaluating}.
The methodological lesson is not that shuffled contexts are universally the
correct control for contextual NLP tasks. Rather, the appropriate control
should target the hypothesized source of information while making clear what
the system is and is not required to infer. In MHER, the shuffled condition
provides a transparent stress test of system behavior when pair-specific
grounding is deliberately invalidated and disclosed. A future blinded variant,
in which models receive misaligned evidence without being told that alignment
has been broken, would test the distinct capability of independently detecting
provenance inconsistency.

\subsection{Model Policy Should Not Be Confused with Historical Capability}
\label{sec:discussion-model-policy}

Another finding is that the evaluated systems differ sharply in how they
respond to limited or misleading evidence. The first-party API models
frequently abstain under Name-only input, whereas Qwen3-8B resolves every
Name-only item. Under the explicitly signaled shuffled condition, some systems
again respond primarily through abstention, while others continue to make
binary decisions.

These differences caution against interpreting raw benchmark accuracy without
coverage or uncertainty behavior. A low ordinary accuracy may arise because a
model is wrong, because it refuses to decide, or because its decision policy
is deliberately conservative. Conversely, a model can achieve full coverage
by making many low-quality binary resolutions.

This is why \textsc{ambiguous} is treated in MHER as a selective-prediction
action rather than as a third historical relation. Selective prediction
separates the decision to answer from the correctness of the answer
\citep{xin2021art,wen2025know}, and the present results show that this
distinction is especially important when comparing models whose refusal
policies differ substantially.

The expert-unresolved challenge provides a narrower test of this interface.
Its condition-specific instruction explicitly informs models that the ten
pairs come from scholarship in which the identity question is not securely
resolved and reminds them that a binary answer is not required. Under this
explicitly signaled unresolved condition, all five systems choose
\textsc{ambiguous} for all ten cases.

This result should therefore be interpreted as \emph{successful use of the
available abstention action under an expert-designated unresolved condition},
not as evidence that the models independently inferred historiographical
uncertainty from the names themselves. Nor does it establish how they would
behave if uncertainty had to be diagnosed without an explicit condition cue.
A stronger future challenge could withhold the unresolved designation while
supplying neutral, non-leaking historical evidence and test whether models can
identify insufficient evidence autonomously.

The model matrix also illustrates why provenance of the evaluated system
matters. The formal comparison combines first-party API models with an exact,
hash-pinned open-weight checkpoint rather than relying on nominal model names
alone. The substantial behavioral differences between these systems show that
Name-only performance is strongly shaped by model-specific decision policy,
particularly the propensity to abstain. For reproducible historical NLP, reporting the source of the model,
its exact checkpoint when available, and the inference configuration is
therefore part of interpreting the result rather than merely an engineering
detail.

At the same time, MHER should not be treated primarily as a model leaderboard.
The five systems differ in size, provider, deployment environment, and
abstention policy. Their value in the present study is that they provide
heterogeneous probes of the same historical reconciliation framework. The
most stable finding is not which system is ``best,'' but that correctly
source-grounded historical evidence supports highly reliable identity
resolution across systems whose Name-only behavior differs dramatically.
The explicit misgrounding controls further show that these systems respond
very differently when the validity of the supplied evidence is deliberately
broken and disclosed.

\subsection{Toward Evidence-Centered Historical NLP}
\label{sec:discussion-future}

Taken together, the results suggest an evidence-centered direction for
historical NLP.

Many historical NLP pipelines can be organized conceptually as a progression
from document processing to entity extraction, entity linking, and structured
historical knowledge. MHER focuses on a step that often remains implicit
within that progression: determining whether heterogeneous attestations should
be collapsed into one historical person in the first place.

For this step, increasing parametric knowledge is only one possible route.
An alternative is to expose the documentary evidence relevant to the decision
and evaluate how model judgments change across explicitly controlled evidence
conditions. This has several potential advantages. It makes long-tail
identities less dependent on pretraining exposure; it provides a natural
interface for source citation and human verification; it permits task designs
in which abstention is available rather than forcing every identity question
into a binary decision; and it enables benchmark designers to probe how
systems respond when evidence is absent, sufficient, removed, or deliberately
misgrounded.

The present study should therefore be read less as evidence that modern
language models have ``solved'' historical identity resolution and more as a
demonstration that source-grounded evaluation changes what is being measured.
Names alone ask, in part, what a model already knows or is willing to infer
from form. Source-grounded reconciliation asks whether an identity relation
can be reconstructed from evidence visible in the record.

That distinction may be especially important for low-resource and archival
settings. The historical people who matter for research are not restricted to
those who are prominent enough to be well represented in contemporary
knowledge bases or language-model pretraining. An evidence-centered system
can, in principle, work from what survives in the sources rather than from
what survives in model memory.

MHER provides one controlled instance of this broader idea. Extending it will
require other historical traditions, longer and noisier documents, incomplete
or conflicting evidence, additional entity types, closer integration with
historian-facing workflows, and stronger blinded controls in which models must
detect unreliable or insufficient evidence without being explicitly told that
a condition is misgrounded or unresolved.

The central design principle is nevertheless broader: when historical context
is used as computational evidence, evaluation should make its provenance
explicit and distinguish the correctness of the identity judgment from the
conditions under which that judgment was produced.

\section{Limitations}
\label{sec:limitations}

MHER is intentionally designed as a controlled benchmark for one specific
historical identity problem. Its provenance requirements, paired intervention,
and falsification controls strengthen the interpretation of the reported
effects, but they also define important limits on what can be concluded from
the present study.

\paragraph{Domain and task scope.}
MHER focuses on personal names associated with the Mongol world and therefore
does not establish that the same results will generalize to other historical
periods, regions, documentary traditions, or entity types. Naming systems,
transcription practices, source density, and the availability of corroborating
evidence differ substantially across historical domains.

The current benchmark also begins from already identified person-name
mentions. It does not evaluate named-entity recognition, candidate retrieval,
document retrieval, or the end-to-end discovery of relevant historical
evidence. Nor does it test organizations, places, titles, events, or other
entity types for which the evidential structure of reconciliation may differ.
The present contribution should therefore be understood as a controlled study
of person-level identity reconciliation rather than a complete historical
entity-linking pipeline.

\paragraph{Evidence-eligible selection and near-ceiling performance.}
The Source-grounded condition is a deliberately stricter subset of the
Name-only benchmark. An item enters this subset only when both mentions can be
associated with provenance-traceable evidence and, for \textsc{same} pairs,
when the two sides satisfy the source-independence requirements described in
Section~\ref{sec:source-grounded-construction}. The 108-item paired TEST set
therefore represents cases for which sufficiently informative source evidence
could be assembled under the benchmark protocol.

This selection is important for interpreting the near-ceiling performance of
the first-party systems. The results show that the \emph{selected,
provenance-complete evidence} is highly sufficient for reconciliation; they do
not imply that equally informative evidence will always exist in historical
archives. Sparse, damaged, contradictory, poorly indexed, or unevenly
preserved sources may provide much weaker support.

The current contexts are also deliberately compact and evidence-focused.
Future evaluation should include longer source passages, partially relevant
documents, conflicting testimony, and settings in which relevant evidence must
first be retrieved from a larger archive. Such extensions would test not only
whether a model can use sufficient evidence once exposed to it, but whether it
can identify and combine useful evidence under more realistic archival
conditions.

Near-ceiling Source-grounded performance also limits fine-grained comparison
among the strongest systems. MHER is therefore better suited to studying the
effect of evidence conditions than to ranking near-ceiling models under the
Source-grounded condition itself.

\paragraph{Gold curation and scholarly uncertainty.}
Gold identity labels are derived from a curated canonical registry and
curator-facing scholarly evidence records rather than from model agreement.
All 84 primary entities underwent a systematic curator re-review against their
registered scholarly evidence. This provides explicit provenance for the
binary gold relation, but it is not equivalent to independent double
annotation by multiple historians.

Historical identity judgments can themselves be contested, particularly for
poorly documented or philologically ambiguous individuals. We therefore
exclude unresolved scholarly identifications from the primary binary core
rather than forcing them into \textsc{same} or \textsc{different}. The
separate MHER-Unresolved challenge contains only ten cases and should be
interpreted descriptively rather than as a statistically comprehensive
benchmark of historiographical uncertainty.

Moreover, the unresolved challenge is currently Name-only. We do not evaluate
a Source-grounded unresolved condition because the available scholarly
uncertainty notes contain explicit uncertainty or equivalence language that
would reveal the desired abstention behavior. A stronger future design would
require independently sourced, neutral evidence for unresolved cases that can
be shown not to leak either the uncertainty judgment or a preferred
reconciliation.

\paragraph{Benchmark size and repeated entities.}
The primary benchmark contains 84 historical persons and 396 Name-only pairs,
with 108 TEST pairs eligible for the central Source-grounded intervention.
This is smaller than large-scale contemporary entity-linking benchmarks.
The size reflects the cost of constructing source-linked, manually audited
historical evidence rather than an attempt to maximize pair count.

Multiple pairs within a split may involve the same canonical person.
Entity-disjoint DEV/TEST splitting prevents identity leakage across
development and test partitions, and the repeated-entity analyses in
Section~\ref{sec:repeated-entity-results} show that the principal effects are
not driven by one or a few heavily represented entities. Nevertheless, a
larger registry spanning more people and more independent historical
communities would provide stronger estimates of cross-entity and
cross-domain generalization.

\paragraph{Prominence is only a proxy for model familiarity.}
The \textsc{head}, \textsc{mid}, and \textsc{long-tail} strata are assigned
before evaluation and provide a useful way to examine whether Name-only
behavior varies with public visibility. They are not measurements of actual
training-data frequency or proof that a model has or has not encountered a
particular historical person during pretraining.

Accordingly, the prominence analyses in
Section~\ref{sec:prominence-results} should be interpreted as sensitivity to
a coarse visibility proxy rather than as direct evidence of memorization.
Model-specific abstention policies, orthographic regularity, and differential
coverage of names and languages can all contribute to the observed
associations. Establishing actual training exposure would require access to
training corpora or substantially different experimental designs.

\paragraph{Model identity, deployment, and reproducibility.}
The formal model matrix includes first-party API systems and one frozen
local open-weight deployment. Commercial API systems remain versioned services
whose internal weights, decoding implementation, or safety policies may change
over time, and the local Qwen3-8B result characterizes one fixed deployment
rather than every possible implementation of that model family.

These considerations are especially relevant to Name-only evaluation, where
the formal systems exhibit sharply different abstention and decision policies.
MHER consequently treats model provenance and inference configuration as part
of result interpretation rather than as incidental implementation metadata.
Exact deployment records are preserved in the frozen project state but are
not released in this arXiv version.

\paragraph{Source representation, copyright, and release.}
MHER relies on historical and modern scholarly sources whose legal and
practical redistribution conditions differ. Upon release, MHER will distribute project-authored evidence paraphrases
with bibliographic citations and source locators rather than scans or
extended passages from copyrighted modern editions.

This policy supports source auditing while reducing redistribution risk, but
it also means that the released contexts will be curated representations of
the underlying source evidence rather than substitutes for the original
documents. Researchers seeking to verify an interpretation at the level of the
source text must consult the cited edition or archival object itself.

More generally, provenance control does not eliminate interpretive judgment
from historical data construction. Decisions about what constitutes a neutral
paraphrase, which source supports a mention, and when two descriptions provide
independent evidence remain scholarly choices. MHER makes these choices
explicit and auditable, but it cannot make them entirely mechanical.

\paragraph{Summary of scope.}
The strongest conclusion supported by MHER is therefore deliberately narrow.
For this frozen Mongol-world benchmark, and for identity questions for which
independently provenance-controlled evidence can be supplied, correctly
grounded historical evidence produces substantially more reliable
reconciliation than name forms alone or mismatched evidence across diverse
model families.

The present experiments do not establish that historical entity
reconciliation is solved in general, that source evidence will always be
available or sufficient, or that the observed behavior transfers unchanged to
other historical traditions. Those questions require broader benchmarks with
more heterogeneous archives, less complete evidence, and independent
historical validation.

\section{Conclusion}
\label{sec:conclusion}

We introduced MHER, a provenance-controlled benchmark for historical entity
reconciliation across languages, scripts, and transcription traditions in
sources concerning the Mongol world. Rather than treating historical identity
as a problem of surface-name correspondence or requiring every mention to
resolve first to a modern knowledge-base entity, MHER asks directly whether
two source-attested person mentions refer to the same historical individual.

The experiments reveal a consistent distinction between names and evidence.
Name-only reconciliation is strongly model-policy dependent: some systems
respond to uncertain names primarily through abstention, whereas others make
aggressive binary decisions. When correctly source-grounded historical
evidence is supplied, these differences contract substantially and identity
resolution becomes markedly more reliable across model families. The
identical-surface hard negatives make this distinction especially clear:
cases that cannot be separated from the visible names alone become resolvable
once chronology, kinship, office, place, and other historical evidence are
made available.

At the same time, the ablations show that the effect is not simply a
consequence of supplying more text. Context alone carries substantial identity information, while the same
context pool produces much lower performance under the explicitly signaled
misgrounding controls than under correct Source-grounded evidence. These
results support treating provenance as part of the benchmark semantics,
while the present controls do not isolate evidence alignment from the
condition instruction itself. The Qwen3-8B analysis further
shows that names are not uniformly beneficial once strong context is
available: surface forms can sometimes interfere with otherwise sufficient
historical evidence and induce false identity merges.

These findings suggest a broader direction for historical NLP. Historical
identity should not be evaluated solely by asking whether a model already
recognizes a person from a name. A more informative question is whether the
model's judgment changes appropriately with the evidence preserved in the
historical record. MHER provides one controlled setting for studying this
problem and, more generally, illustrates why source provenance, abstention,
and falsification controls should be treated as part of the evaluation design
when historical context itself is used as evidence.

Names provide evidence about identity; they are not themselves the identity
relation.

\bibliographystyle{plainnat}
\bibliography{references}

\end{document}